\documentclass[sigconf]{acmart}

\setcopyright{none} 
\renewcommand\footnotetextcopyrightpermission[1]{}
\usepackage{tikz}
\usepackage{amsmath}
\usepackage{listings, lstautogobble}
\usepackage{booktabs}
\usepackage{subcaption}
\usepackage{diagbox}
\usepackage{url}
\usepackage{xcolor}
\usepackage{xspace}

\usepackage{multirow}
\usepackage{array}
\newcolumntype{P}[1]{>{\centering\arraybackslash}p{#1}}
\newcolumntype{M}[1]{>{\centering\arraybackslash}m{#1}}

\definecolor{comment_color}{HTML}{bf0f02}
\definecolor{darkred}{HTML}{b02300}

\newcommand{\sysname}{ExAD}

\makeatletter
\g@addto@macro{\UrlBreaks}{\UrlOrds}
\makeatother

\lstdefinestyle{mystyle}{
 basicstyle={\ttfamily\scriptsize\linespread{0}},
 commentstyle=\color{gray}\ttfamily,
 captionpos=b,
 frame=single,
 showstringspaces=false,
 numberstyle=\color{gray},
 numbers=left,
 numbersep=5pt,
 xleftmargin=10pt,
}
\makeatletter
\def\maxwidth{%
  \ifdim\Gin@nat@width>\linewidth
    \linewidth
  \else
    \Gin@nat@width
  \fi
}
\makeatother

\AtBeginDocument{%
  \providecommand\BibTeX{{%
    \normalfont B\kern-0.5em{\scshape i\kern-0.25em b}\kern-0.8em\TeX}}}

\begin{document}

\title{ExAD: An Ensemble Approach for Explanation-based Adversarial Detection}

\author{Raj Vardhan}
\affiliation{
\institution{Texas A\&M University}
}
\email{raj_vardhan@tamu.edu}

\author{Ninghao Liu}
\affiliation{
\institution{Texas A\&M University}
}
\email{nhliu43@tamu.edu}

\author{Phakpoom Chinprutthiwong}
\affiliation{
\institution{Texas A\&M University}
}
\email{cpx0rpc@tamu.edu}

\author{Weijie Fu}
\affiliation{
\institution{Hefei University of Technology}
}
\email{fwj.edu@gmail.com}

\author{Zhenyu Hu}
\affiliation{
\institution{Texas A\&M University}
}
\email{johnhu@tamu.edu}

\author{Xia Ben Hu}
\affiliation{
\institution{Texas A\&M University}
}
\email{hu@cse.tamu.edu}

\author{Guofei Gu}
\affiliation{
\institution{Texas A\&M University}
}
\email{guofei@cse.tamu.edu}

\begin{abstract}
Recent research has shown Deep Neural Networks (DNNs) to be vulnerable to adversarial examples that induce desired misclassifications in the models. Such risks impede the application of machine learning in security-sensitive domains. Several defense methods have been proposed against adversarial attacks to detect adversarial examples at test time or to make machine learning models more robust. However, while existing methods are quite effective under blackbox threat model, where the attacker is not aware of the defense, they are relatively ineffective under whitebox threat model, where the attacker has full knowledge of the defense.  

In this paper, we propose ExAD, a framework to detect adversarial examples using an ensemble of explanation techniques. Each explanation technique in ExAD produces an explanation map identifying the relevance of input variables for the model's classification. For every class in a dataset, the system includes a detector network, corresponding to each explanation technique, which is trained to distinguish between normal and abnormal explanation maps. At test time, if the explanation map of an input is detected as abnormal by any detector model of the classified class, then we consider the input to be an adversarial example. We evaluate our approach using six state-of-the-art adversarial attacks on three image datasets. Our extensive evaluation shows that our mechanism can effectively detect these attacks under blackbox threat model with limited false-positives.
Furthermore, we find that our approach achieves promising results in limiting the success rate of whitebox attacks.

\end{abstract}

\maketitle
\pagestyle{plain}
\section{Introduction}
\begin{figure}[t]
  \centering
  \includegraphics[width=0.4\linewidth]{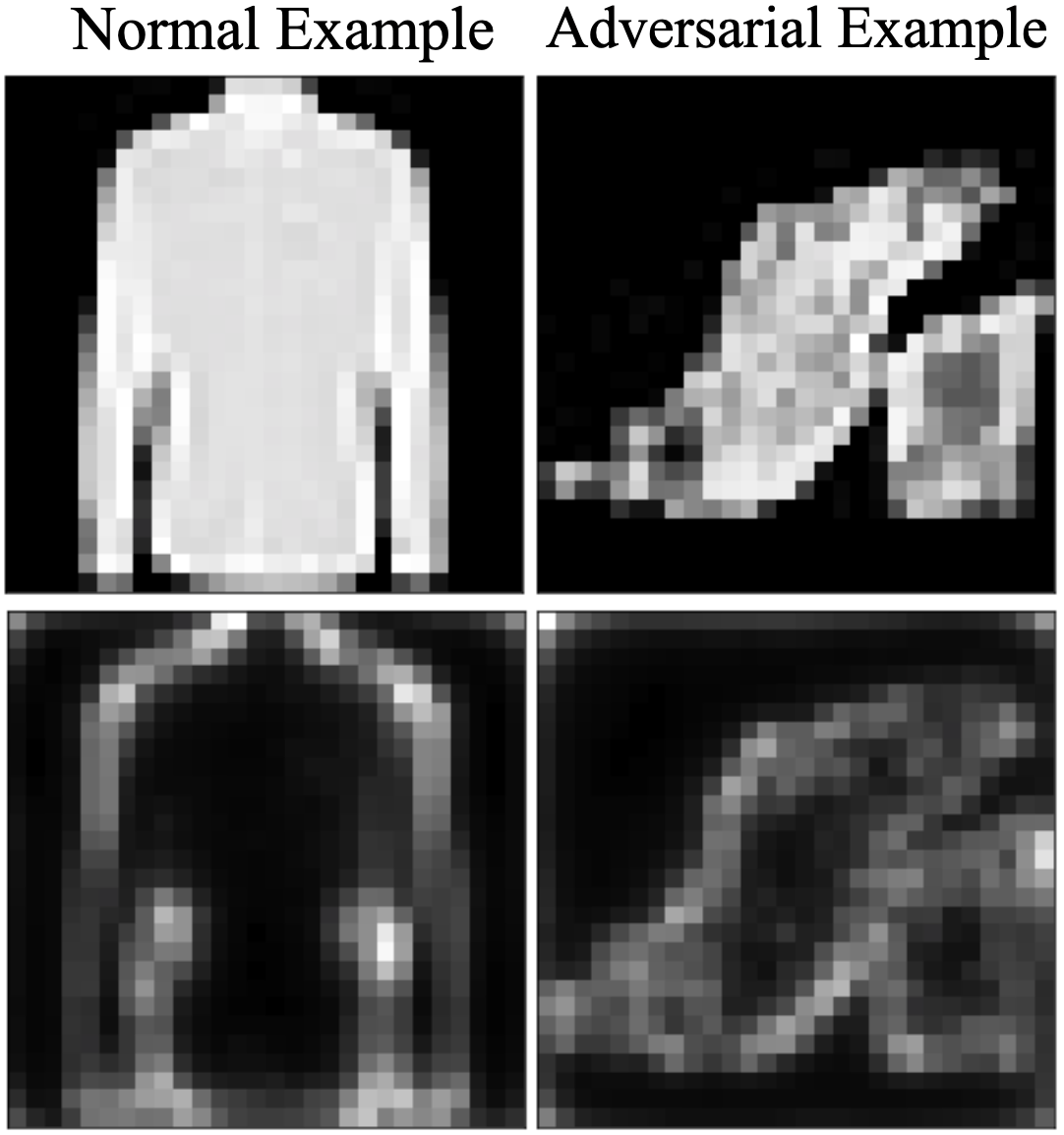}
   \caption{Intuition behind the proposed \sysname{} framework. 
   }
  \label{fig:intuition}
  \vspace{-3mm}
\end{figure}
In recent years, Deep Neural Networks (DNNs) are being increasingly adopted in a wide range of tasks such as face-recognition \cite{krizhevsky2012imagenet}, natural language processing \cite{hinton2012deep}, and malware classification \cite{dahl2013large}. This trend can be attributed to the superior performance achieved by DNNs in solving computational tasks that rely on high-dimensional data. However, increasing adoption of DNNs to security-critical applications, such as self-driving cars and malware classification, is hindered by the vulnerability of DNNs to adversarial attacks \cite{szegedy2013intriguing, papernot2016limitations, goodfellow2014explaining, liu2016delving}. Specifically, minor yet carefully computed perturbations to natural inputs can cause DNNs to misclassify.

Several methods have been proposed for defending against adversarial examples. One direction of research is to improve the robustness of neural networks, such as through adversarial training \cite{goodfellow2014explaining} or gradient masking \cite{gu2014towards}. However, subsequent works have shown that neural network architectures modified with such techniques can still be attacked \cite{carlini2017towards}. Another research direction is adversarial detection, where the goal is to detect if an input is an adversarial example or a normal example. Early works in this area either used a second neural network \cite{Gong-etal17notTwins,Grosse-etal17statisticDetection,Metzen-etal17detectPerturbation}, or statistical tests \cite{hendrycks2016early,li2017adversarial,bhagoji2017dimensionality} to classify between normal and adversarial examples. However, Carlini et al.~\cite{Carlini-Wagner17notEasilyDetected} showed that while most of these mechanisms are successful against blackbox attacks, they lack robustness to whitebox attacks, where the adversary has knowledge of the defense. 
Although many recent methods have enhanced the detection of blackbox adversarial attacks \cite{magnet, Ma-etal19NIC, Xu-etal17feature, ma2018characterizing}, improving the robustness to whitebox attacks remains an open problem. 

One way of uncovering the reasons for the resulting misclassification of an adversarial example can be understanding why the model predicts what it predicts through explanation techniques \cite{Simonyan-etal13deepInsideCNNsaliency, baehrens2009explain,shrikumar2016not,Sundararajan-etal16integratedGradient,springenberg2014striving,Bach-etal15onPixelwise, montavon2017explaining,kindermans2017learning}. For an image input, the result from an explanation technique encodes the relevance of each pixel for the prediction result and is commonly referred to as an \textit{explanation map}.
Our hypothesis is that the explanation map of an adversarial example being misclassified as the target class may not be consistent with explanation maps generated for correctly classified normal examples of that class. We term the former type as \textit{abnormal explanations} and the latter as \textit{normal explanations} throughout this paper. 
Figure \ref{fig:intuition} shows an intuitive example where we can observe that the explanation map of an adversarial example classified into the shirt class (bottom-right) is quite distinguishable from that of a normal example of the targeted class (bottom-left).
Overall, the distinguishability between normal and abnormal explanation maps guides us in exploring the effectiveness of using explainability as a tool for detecting adversarial examples.

However, a defense method that relies on a single explanation technique may still not be robust under whitebox setting. An adaptive adversary can leverage recent findings which show that explanations can be unreliable \cite{kindermans2017reliability} and can be manipulated to produce a target explanation map \cite{zhang2020interpretable, dombrowski2019explanations}. Such an adaptive adversary can generate adversarial examples that not only fool the target model into producing desired misclassifications, but also fool the targeted explanation technique into producing normal explanation maps.
Towards building a mitigation strategy, we take motivation from previous work on N-variant systems~\cite{cox2006n}. To provide higher resistance against attacks on software vulnerabilities, these systems combine multiple variants with disjoint exploitation sets into a single system.
In context of our work, we propose to use an ensemble of multiple kinds of explanation techniques. The benefit of this approach is that it requires an adaptive adversary to construct an adversarial example that fools the target model and simultaneously fools all explanation techniques. By incorporating diverse explanation techniques, we can reduce the probability that an attacker will achieve this goal. 

In this work, using the above insights, we propose \sysname, an \underline{E}nsemble approach for E\underline{x}planation-based \underline{A}dversarial \underline{D}etection. 
\sysname{} uses an ensemble of explanation techniques wherein each technique provides an explanation map for every classification decision by a target model. 
To introduce explanation diversity, we include both gradient-based~\cite{Simonyan-etal13deepInsideCNNsaliency, baehrens2009explain,Smilkov-etal18smoothgrad,Sundararajan-etal16integratedGradient,shrikumar2016not} and propagation-based~\cite{Bach-etal15onPixelwise, montavon2017explaining, shrikumar2016not} explanation techniques in \sysname.
Furthermore, for any class in a dataset, the system includes a detector model associated with each explanation technique. The detector model determines if an explanation map produced by the respective explanation technique is normal or not for that class. The key idea here is to use the distinguishability between normal and abnormal explanations for any class.
Finally, for a test input classified into a particular class, if the explanation map produced by any technique is detected as abnormal by the corresponding detector model, then we classify the input as an adversarial example. 

We evaluate \sysname{} using six state-of-the-art adversarial attacks on three image datasets, namely MNIST \cite{backprop}, Fashion-MNIST (FMNIST) \cite{fmnist} and CIFAR-10 \cite{krizhevsky2009learning}. We first perform the evaluation under the blackbox threat model. Our experimental results show that we can effectively detect all attacks, achieving a detection rate above 98\% (many having 100\% detection rate) across the three datasets with a low false-positive rate of under 1.1\%. 

More importantly, we further evaluate \sysname{} under whitebox threat model. We build on previous research~\cite{dombrowski2019explanations, zhang2020interpretable}, and create a strong adaptive adversary to generate adversarial examples that fool the target model as well as a target explanation technique.
Through experimental results, we make an interesting finding on the transferability of adaptive attacks on explanation-based detector models. We observe that on targeting a propagation-based technique, the resulting adversarial examples are more successful in fooling detector models of other propagation-based techniques (into misclassifying an explanation map as normal) as compared to fooling detector models of gradient-based techniques. 
Likewise, we find that targeting a gradient-based technique transfers better to the detector model of the other gradient-based technique compared to those of propagation-based techniques. 
Using an ensemble of detector models corresponding to diverse techniques, \sysname{} achieves a mean detection rate of over 88\% for this whitebox attack across the three datasets. The results indicate that our proposed defense can significantly limit the success rate of such whitebox attacks. Additionally, we find that our ensemble approach makes it considerably harder for attackers to perform more advanced whitebox attacks, such as simultaneously targeting all explanation techniques.

We summarize our main contributions as follows.
\begin{itemize}
    \item We develop a novel framework called \sysname{} to detect adversarial examples. \sysname{} uses an ensemble of diverse explanation techniques to improve the robustness against whitebox attacks.

    \item We evaluate \sysname{} on six state-of-the-art adversarial attacks and three image datasets under blackbox threat model. The results show that the proposed system can consistently achieve high detection rates with a low false-positive rate.

    \item We extensively evaluate \sysname{} under whitebox threat model by creating a strong adaptive adversary which targets the classification model as well as an explanation technique. Our findings show that \sysname{} achieves promising results in limiting the success rate of whitebox attacks. 

\end{itemize}

The rest of the paper is organized as follows. In Section \ref{sec_background}, we review background and related work. In Section \ref{sec_design}, we introduce our proposed framework in light of two applicable threat models. Subsequently, we report experimental results and comparison with state-of-the-art detection methods in Section \ref{sec_experiment}. Then, we discuss aspects such as the fragility of explanations and limitations of our work in Section \ref{discussion}. Finally, we conclude the paper in Section \ref{sec_conclusion}.
\section{Background and Related Work}
\label{sec_background}

\subsection{Neural networks}
A DNN is a computational graph of elementary computing units, called neurons, organized into layers that represent the extraction of successive representations from the input. We use notations consistent with previous work \cite{Ma-etal19NIC,Carlini-Wagner17notEasilyDetected} to denote an $m$-class DNN as a function $f : \Bbb R^d \rightarrow \Bbb R^m$. 
The i-th layer of the network computes
\begin{equation*}
f^i(x) = ReLU(W^i f^{i-1}(x) + b^i)
\end{equation*}
where $W^i$ is a weight matrix, $b^i$ is a vector of bias values, and ReLU is a non-linear activation function. Let $Z(x)$ denote a vector of $m$ elements representing the output of the last layer (before softmax), known as \textit{logits}, i.e., $Z(x) = f^n(x)$. A softmax function is used to obtain the normalized output of the network given by  $y = f(x) = \textnormal{softmax}(Z(x))$ where $x \in \Bbb R^d$ and $y \in \Bbb R^m$ with $y_i$ representing the probability of the input being recognized as class $i$. Then, we represent the classification of $f(\cdot)$ on $x$ by $C(x) = \textnormal{argmax}_i(f(x)_i)$.
At test-time, a trained model is provided with test inputs $X_t$, and for each input $x_t \in X_t$, the model assigns its classification to be $C(x_t) = argmax_i(f(x_t)_i)$. 
The classification is considered correct if $C(x_t)$ is same as the true label $C^*(x_t)$.

\subsection{Adversarial examples}
\label{background_adversarial_examples}
Adversarial examples are crafted by imperceptibly perturbing normal inputs to cause DNNs into misclassifying them. Formally, an input to the classifier $f(\cdot)$ is termed as normal if it occurs naturally \cite{magnet} or was benignly created \cite{Carlini-Wagner17notEasilyDetected}. Then, given a normal input $x \in \Bbb R^d$ with correctly classified class $C(x) = c$, we call $x'$ an (\textit{untargeted}) adversarial example if it is close to $x$, i.e., $\Delta(x, x') < \epsilon$ and $C(x') \neq c$, where $\Delta(.)$ denotes a measure of similarity between two inputs and $\epsilon$ is a threshold that limits the permissible perturbations in the adversarial example. In a more restrictive case, an attacker could also target a desired class $t \neq c$ and find a $x'$ close to $x$ such that $C(x) = c$ and $C(x') = t$. We call $x'$ a \textit{targeted} adversarial example.

In the case of images, the closeness function $\Delta(.)$ and threshold $\epsilon$ should be chosen such that the adversarial example and its \textit{seed image} (normal counterpart) are indistinguishable to a human eye. To define $\Delta(.)$, a popular distance metric is the $L_p$ norm, defined as $\|d\|_p = \Big(\sum^{n}_{i=0}|v_i|^p\Big)^{\frac{1}{p}}$. Common choices for $L_p$ include: $L_0$, a measure of the number of pixels which have different values in corresponding positions in two images; $L_2$, which measures the standard Euclidean distance; or $L_{\infty}$, a measure of the maximum change among all pixels at corresponding places in two images.

\subsection{Existing attacks}
\label{subsec_existing_attacks}
Researchers have developed a number of methods for constructing adversarial examples. Broadly, these methods can be categorized into \textit{gradient-based} attacks \cite{carlini2017towards, goodfellow2014explaining, szegedy2013intriguing}, which leverage gradient-based optimizations, and \textit{content-based} attacks \cite{brown2017adversarial, eykholt2017robust}, where perturbations are made in accordance with the semantics of the input content to simulate real-world scenarios. In this paper, we focus on six state-of-the-art gradient-based attacks for neural network classifiers, namely Jacobian-based Saliency Map Attack (JSMA) \cite{papernot2016limitations}, Basic Iterative Method (BIM) \cite{bim}, Momentum Iterative Method (MIM) \cite{dong2018boosting}, and Carlini and Wagner Attacks (CW) \cite{carlini2017towards} tailored to $L_0$, $L_2$, and $L_{\infty}$ norms. For more details, we refer interested readers to the original papers and to recent works on adversarial detection \cite{Ma-etal19NIC, magnet} which provide a good summary of these attacks.

\subsection{Existing work on adversarial detection}
Adversarial detection is a defense approach with the goal of building a classifier $g$ with a binary output $y\in \{0, 1\}$, where labels $0$ and $1$ denote that the input instance is normal or adversarial, respectively. 
We briefly review state-of-the-art works in detecting adversarial examples, and divide them into three categories as below.

\textbf{Training a Detector.} First, we can use adversarial examples to train detectors. The input into detectors can be chosen as data instances in raw feature space or the intermediate representation space of the target model. Using the former strategy, Gong~\textit{et al.} show that a simple binary classifier can learn to separate normal and adversarial instances~\cite{Gong-etal17notTwins}. 
In a related work, Grosse et al. add a new class, solely for adversarial examples, in the output layer of the model~\cite{Grosse-etal17statisticDetection}. 
But, modifying the model architecture impacts the accuracy on normal examples. Based on the latter approach, Metzen~\textit{et al.} use representations generated by inner deep neural network layers as inputs into detectors which are augmented to the classification network~\cite{Metzen-etal17detectPerturbation}. By freezing the weights of the classification network before training the detectors, this method does not affect the classification accuracy on normal examples. However, in subsequent work, Carlini et al. showed that these detectors don't generalize well and lack robustness to whitebox attacks~\cite{Carlini-Wagner17notEasilyDetected}. In our work, we mitigate the generalization challenge by including an attack-independent defense setting (discussed in Section \ref{subsec_detection_ae}).



\textbf{Statistical Metrics.} Second, we can use statistical metrics to design detectors. Grosse \textit{et al.}~\cite{Grosse-etal17statisticDetection} study two statistical distance measures, Maximum-Mean-Discrepancy and Energy Distance, where a sample is regarded as adversarial if it is rejected by statistical testing. 
Ma et al. estimate an LID value which assesses the space-filling capability of the region around an example by measuring the distance distribution with respect to its neighbors~\cite{ma2018characterizing}. The authors demonstrate that estimated LID of adversarial examples tends to be much higher than that of normal examples.
However, a challenge faced by these approaches is in developing more effective and transferable metrics to separate clean instances from adversarial examples generated by different attacks.

\textbf{Prediction Abnormality.} Third, we can also resort to detecting the abnormality of input instances. 
Meng and Chen proposed Magnet~\cite{magnet} which learns to approximate the manifold of normal examples using autoencoders.
Another method called Feature Squeezing \cite{Xu-etal17feature} proposes reducing the degree of freedom of an adversary, such as by smoothing images or minimizing their color depth.
Another recent work called Neural-network Invariant Checking (NIC) proposed leveraging the provenance channel and the activation value distribution channel in DNNs by showing that adversarial examples tend to violate either provenance invariant or value invariant~\cite{Ma-etal19NIC}.
However, while these methods have improved the detection rates on blackbox attacks, they have shown very limited success against whitebox attacks. 
Zhang~\textit{et al.} proposed a detection method based on perturbation of saliency maps~\cite{zhang2018detecting}. 
The authors find that on adding adversarial perturbations, the saliency of the adversarial example is also perturbed compared to that of the seed image. However, this difference between saliency may not be effective because, at test-time, we do not know the class from which an adversarial example originated. Therefore, we do not know what its normal saliency would look like had the example not been perturbed. 
Besides, explanations have also been used by Liu et al. for a different goal of crafting adversarial examples ~\cite{liu2018adversarial}.
In this paper, we further explore the premise of using explanations, but to detect adversarial examples and based on a fundamentally different approach. Our work was done concurrently with a similar approach presented by Wang et al.~\cite{wang2020interpretability}. 
In contrast, we have the following differentiating aspects. First, our work offers a more detailed evaluation on whitebox attacks. Second, we provide a discussion on the fragility of an explanations-based defense. Finally, we compare the proposed method with a number of state-of-the-art detection systems.

\vspace{-3mm}
\subsection{Explainable machine learning}
\label{background:intr_ml}

Our work utilizes recent advances in explainable machine learning. Specifically, we focus on \textit{local explainability} methods~\cite{baehrens2009explain,lipton2016mythos} which explain the output of DNN models for a given input. For computer vision models, these techniques identify which regions in an input image are most responsible for the prediction result. The explanation result is often termed as a \textit{saliency map}~\cite{Simonyan-etal13deepInsideCNNsaliency}, or more generally, an \textit{explanation map}~\cite{dombrowski2019explanations}. Naturally, our defense is compatible with models that are inherently explainable (e.g., linear models) and models that produce an explanation result along with the prediction~\cite{Rudin19pleaseStop, Luong-etal15attentionTranslation}. However, we focus on local explainability methods as they build on top of existing models. This allows us to add our adversarial detection capability to any existing blackbox model without sacrificing its prediction power for explainability, or putting the burden of producing explanations during classification. 

Among local explanation techniques, backpropagation-based methods have gained considerable attention. These can be further categorized into the following. The first is \textit{gradient-based} techniques which rely on the gradient of the neural network function to generate explanations~\cite{Simonyan-etal13deepInsideCNNsaliency, baehrens2009explain,Smilkov-etal18smoothgrad,Sundararajan-etal16integratedGradient,shrikumar2016not}. The second category is \textit{propagation-based} techniques~\cite{Bach-etal15onPixelwise, montavon2017explaining, shrikumar2016not}. These techniques view the neural network as a computational graph, and generate explanations by starting with the prediction score at the output layer and progressively redistributing it backwards by means of propagation rules until the input layer is reached. To achieve diversity in explanation methods, we use both gradient-based and propagation-based explanation techniques, which we discuss further in Section \ref{gen_of_interpretations}.
\section{Design}
\label{sec_design}

\subsection{Threat model}
\label{sec_threat_model}
In designing our defense, we assume that the attacker has complete knowledge of the target classifier $f(\cdot)$ including its architecture and parameters. This is a conservative and practical assumption, consistent with prior works~\cite{Ma-etal19NIC,Xu-etal17feature, magnet}. Also, depending upon whether the attacker has knowledge of the defense, we consider two types of threat models. First, we consider \textit{blackbox} attack, where the attacker does not have any knowledge of the defense mechanism. Second, we consider \textit{whitebox} attack, where the attacker has complete knowledge of the defense mechanism including its structure and parameters. For \sysname, this implies that the attacker has full knowledge of the explanation techniques and detector models.

\subsection{Overview of \sysname}
\sysname{} is a framework that uses an ensemble of explanation techniques to detect adversarial examples.
The role of explanation techniques is to allow \sysname{} to examine the reasons for the misclassification of adversarial examples. 
Our hypothesis is that the explanations of adversarial examples being misclassified as the target class (\textit{abnormal explanations}) may not be consistent with explanations generated for correct classifications of normal examples of that class (\textit{normal explanations}). 
Our design relies upon the consistency of normal explanations, and their distinguishability from abnormal explanations. 
We provide an intuitive example for the distinguishability aspect in Figure \ref{fig:intuition}. Further examples showing the consistency aspect can be found in Figure \ref{fig:intr_of_gen_and_adv} in Appendix \ref{ap_similarity_normal}. While we provide such motivating examples, it is worth noting that an explanation itself may be incomprehensible to humans as recent work has shown neural networks to use non-robust features (that may not align with human perception of a class) to make predictions~\cite{ilyas2019adversarial}. Therefore, even the distinguishability may not necessarily be apparent to humans. Nevertheless, we empirically show that we can train detector models to learn to distinguish between normal and abnormal explanations.



An overview of our approach is as follows. First, we train the target model as usual on the clean training set. Second, we use a set of diverse explanation techniques to generate explanation maps for normal examples of each class. Third, for every class, we train a detector model corresponding to each technique. The detector model identifies if the explanation map of an example is normal for the classified class. 
We study two approaches to build the detector model: a binary classifier approach (where we use both normal and abnormal explanations) and an anomaly detection approach (where we only use normal explanations). 
Whereas the former setting makes the training and validation process simple (once we have the required data), the motivation for the latter setting is to make our defense attack-independent so that it is more likely to generalize to unknown attacks.
At test time, if the explanation map of an input is classified as abnormal by any detector model of the classified class, then we consider it to be an adversarial example.

\subsection{Generation of explanations}
\label{gen_of_interpretations}

\begin{figure}[t]
  
  \includegraphics[width=\linewidth]{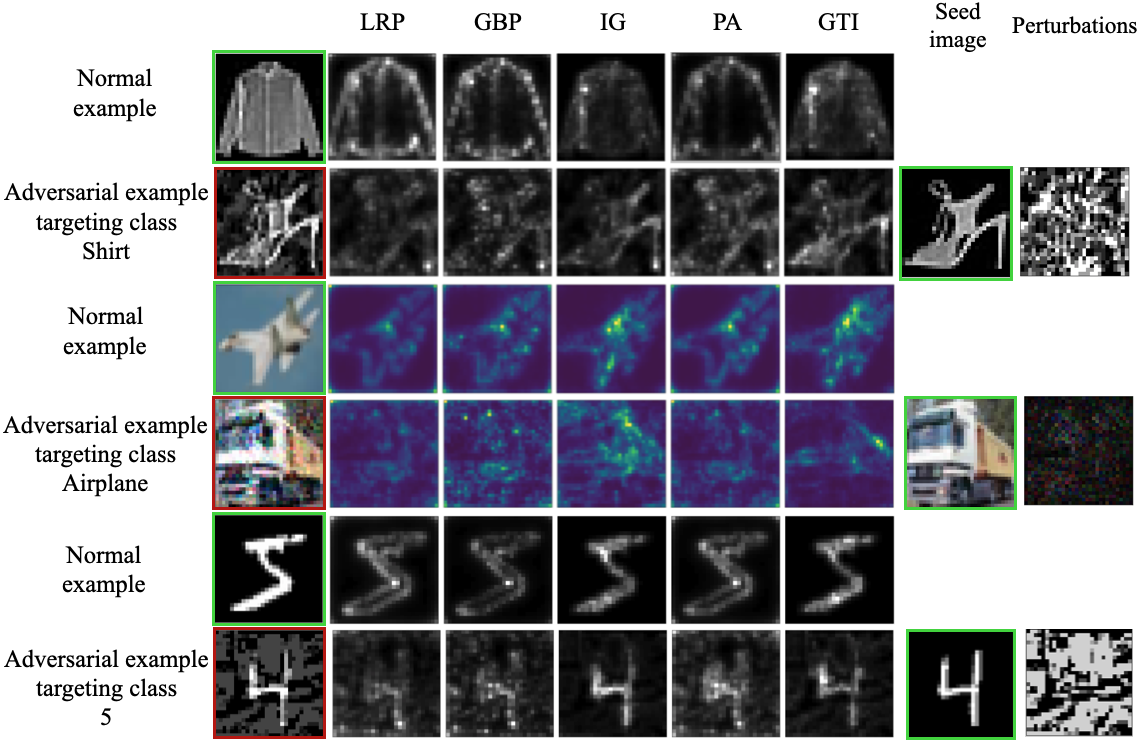}
  \caption{
  \textbf{Distinguishability between normal and abnormal explanations using different explanation techniques.}}
  \label{fig:normal_abnormal}
\vspace{-4mm}
\end{figure}

Given a neural network classifier $f(\cdot)$ and an input $x$, the explanation of the classification of $x$ is represented as an explanation map denoted by $h : \Bbb R^d \rightarrow \Bbb R^d$. The explanation map $h(x)$ encodes the relevance score of every pixel in $x$ for the neural network's prediction. We consider the following explanation generation techniques towards building an ensemble of methods.

\begin{itemize}
    \item \textbf{Gradient}: 
    The gradient of the output $f(x)$ with respect to the input $x$ is indicative of how infinitesimal changes in each pixel can influence the output~\cite{Simonyan-etal13deepInsideCNNsaliency, baehrens2009explain}. The explanation map using the gradient method is given by
    \begin{equation*} 
    h(x) = \frac{\partial f}{\partial x}(x)
    \end{equation*}
    
    \item \textbf{Gradient \textasteriskcentered{} Input (GTI)}: This method computes an element-wise product between the gradient-based explanation map of Simonyan et al. \cite{Simonyan-etal13deepInsideCNNsaliency} and the input to quantify the influence of each pixel on the prediction score \cite{shrikumar2016not}.  
    Formally, the explanation map produced by gradient \textasteriskcentered{} input is given by
    \begin{equation*}
     h(x) = x \odot \frac{\partial f}{\partial x}(x)
    \end{equation*}
    
    \item \textbf{Integrated Gradients (IG)}: In contrast to GTI, which performs a single computation of the gradient at the input $x$, integrated gradients computes the gradients at all points along a linear path from a baseline $\bar{x}$ to $x$, and averages them \cite{Sundararajan-etal16integratedGradient}. The baseline $\bar{x}$ can be defined by the user and is generally chosen as a black image. Formally,
    \begin{equation*}
    h(x) = (x - \bar{x}) \odot \int_{\alpha=0}^1 \frac{\partial f(\bar{x}+\alpha(x-\bar{x}))}{\partial x}\mathrm{d}\alpha
    \end{equation*}
    
    \item \textbf{Guided Backpropagation (GBP)}: This method is an extension of gradient-based explanation with the key difference that it prevents backward flow of negative gradients through non-linearities, such as ReLUs \cite{springenberg2014striving}.
    
    \item \textbf{Layer-wise Relevance propagation (LRP)}: 
    To explain the prediction of class $c$, LRP~\cite{Bach-etal15onPixelwise, montavon2017explaining} starts with the output neuron of class $c$ and goes backwards through the network by following the $z^+$ rule for all layers except the first.
    \begin{equation*}
        R_i^l = \sum_j \frac{x_i^l(W^l)_{ji}^+}{\sum_i x_i^l(W^l)_{ji}^+} R_j^{l+1}
    \end{equation*}
    Here, $i$ and $j$ are two neurons of consecutive layers, $R_i^l$ denotes the relevance of $i$-th neuron in the $l$-th layer, $x_i^l$ represents the activation vector, and $(W^l)_{ji}^+$ denotes the positive weight between the two neurons. Then, to account for the bounded range of an input, we use the $z^\mathcal{B}$ rule in the first layer 
    \begin{equation*}
        R_i^0 = \sum_j \frac{x_j^0 W_{ji}^0 - l_j(W^0)_{ji}^+ - h_j(W^0)_{ji}^-}{\sum_i(x_j^0 W_{ji}^0 - l_j(W^0)_{ji}^+ - h_j(W^0)_{ji}^-)} R_j^1
    \end{equation*}
    where $l$ and $h$ are the lowest and highest allowed pixel values, respectively.
    
    \item \textbf{Pattern Attribution (PA)}: Kindermans et al. \cite{kindermans2017learning} proposed patter attribution as an improvement over the LRP framework. The method is analogous to the backpropagation operation with the weights in the backward pass replaced by element-wise multiplication of weights $W^l$ and learned patterns $A^l$.
    
\end{itemize}

While we considered all six of the above-mentioned explanation techniques to include in \sysname, we found the performance of the gradient method ($h(x) = \frac{\partial f}{\partial x}(x)$) to be unacceptable based on evaluations on the validation sets, whereas remaining techniques performed significantly better. 
Therefore, in this work, \sysname{} uses an ensemble of $k=5$ techniques- LRP, GBP, IG, PA, and GTI. 

Figure \ref{fig:normal_abnormal} shows examples of normal and abnormal explanations produced by different techniques used in \sysname{}. When a test input $x_t$ is classified by the target model $f(\cdot)$ as class $c$, each of the $k$ techniques produce an explanation map for this classification. In column 1, the first, third, and fifth rows show a normal example from FMNIST, CIFAR-10, and MNIST datasets, respectively. In the same column, the second, fourth, and sixth rows show an adversarial example which is misclassified as the class represented by the normal example in the preceding row. Columns 2-6 show the corresponding explanation maps produced by the five explanation techniques. The distinguishability between explanation maps of normal and adversarial examples allows the detector models to determine if an explanation map is normal or not. In the following section, we discuss how a set of $k$ detector models, one corresponding to every explanation technique, evaluate the explanation maps towards determining if $x_t$ is an adversarial example. 


\subsection{Detector models}
\label{subsec_detector}
The detector model determines if the explanation map of an input is normal or abnormal for the classified class. We study the following two methods for building detector models. 

\subsubsection{Detection using a CNN-based binary classifier}
\label{subsec_detection_cnn}
First, we consider an approach of building a CNN-based binary classifier. We term this as the \textit{CNN-based detector model} and denote it as $g(\cdot)$. Under this setting, we refer to the defense as \sysname-CNN. 
Below, we describe the training procedure for these detector models.

For every class $c$, we build a separate detector model $g_{c,j}(\cdot)$ for each explanation technique $h_j$. 
At test-time, for an input classified as class $c$, the detector model $g_{c,j}(\cdot)$ takes the explanation map of the input, produced by the corresponding explanation technique $h_j$, and classifies it as normal or abnormal.
We build this new model $g_{c,j}(\cdot)$ as follows. 
We take every normal example $x_{\textnormal{normal}}$ from class $c$, which is correctly classified by the target model $f(\cdot)$, and generate the explanation map $h_j(x_{\textnormal{normal}})$ for its classification into that class. These explanation maps are considered as normal explanations, and are labeled as negative \textit{class 0}. Then, we generate a number of adversarial examples using different adversarial attacks where the targeted class is $c$. Next, for each successful adversarial example $x_\textnormal{{adv}}$, we generate the explanation map $h_j(x_\textnormal{{adv}})$ for its classification as target class $c$. These explanation maps are considered as abnormal explanations, and labeled as positive \textit{class 1}. Then, we train $g_{c,j}(\cdot)$ on this labeled training set using a CNN-based architecture.


\begin{figure}[t]
  \centering
  \includegraphics[width=0.85\linewidth]{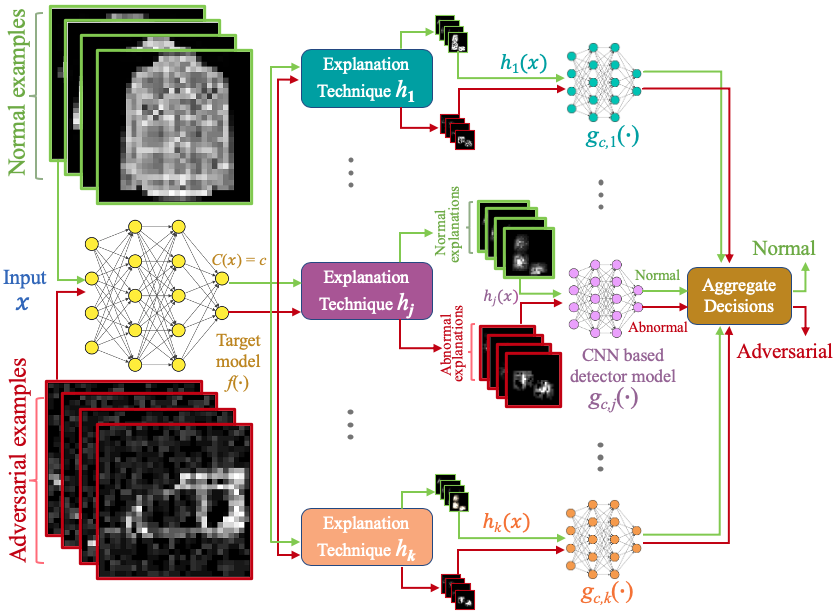}
  \caption{Illustration of the proposed \sysname{} framework}
  \label{fig:workflow}
  \vspace{-3mm}
\end{figure}

\subsubsection{Detection based on reconstruction error}
\label{subsec_detection_ae}
In this approach, we avoid the requirement of adversarial examples to train a detector model, and thereby make the defense more likely to generalize on unknown attacks. Here, we propose using reconstruction error by an autoencoder to determine if the explanation map of a test example is normal or not.
We term this as the \textit{autoencoder-based detector model}. Under this setting, we refer to the defense as \sysname-AE.
Similar to \sysname-CNN setting, we consider each class and build $k$ autoencoder-based detector models, one corresponding to every explanation technique.  



An autoencoder $ae$ = $\psi \circ \phi$ contains two components, an encoder and a decoder, which can be defined as transitions $\phi:\Bbb R^d \rightarrow \Bbb R^z$ and $\psi:\Bbb R^z \rightarrow \Bbb R^d$, respectively,
where $\Bbb R^d$ is the input space and $\Bbb R^z$ is the latent space. For class $c$ and the $j$-th explanation technique, the input space for the autoencoder $ae_{c,j}$ in our system is formed by the set of explanations produced by $h_j$ for correctly classified normal examples of class $c$. We train the autoencoder to minimize a loss function over this set of explanations, where the loss function is taken as the mean squared error (MSE):
\begin{equation*}
    L(\mathbb{E}_{\mathrm{train}}) = \frac{1}{\mathbb{E}_{\mathrm{train}}}\sum_{h_j(x) \in \mathbb{E}_{\mathrm{train}}} \|h_j(x) - (\psi \circ \phi)h_j(x)\|_2
\vspace{-1mm}
\end{equation*}
For a test image, the explanation map $h_j(x)$ produced by the $j$-th technique is given as input to the autoencoder $ae_{c,j}$ which generates a reconstructed image. Then, we compute a reconstruction error:
\begin{equation}
\label{eqn:recon_er}
    R(h_j(x)) = \|h_j(x) - (\psi \circ \phi)h_j(x)\|_p
\end{equation}
where $\|\cdot\|_p$ is a suitable $p$-norm. If the reconstruction error is above a threshold $t_{\textnormal{re}}$, we consider the explanation map $h_j(x)$ to be abnormal. The threshold value is a hyperparameter for each detector model. It should be low enough to detect abnormal explanations, but sufficiently high to not falsely flag normal explanations. We decide $t_{\textnormal{re}}$ values using a validation set of normal explanations, which are in turn derived from a validation set of normal examples. For any detector, we select the highest $t_{\textnormal{re}}$ such that its false-positive rate on the validation set is below a threshold $t_{\textnormal{fp}}$. The threshold $t_{\textnormal{fp}}$ can be chosen depending upon system requirements.

\subsection{Test-time detection of adversarial examples} 
Figure \ref{fig:workflow} illustrates our overall approach, with the \sysname-CNN setting.
At test-time, if an unknown input $x$ is being classified by the target classifier $f(\cdot)$ as class $c$, our goal is to identify if $x$ is a normal example of class $c$ or an adversarial example. To this end, we take the following steps. First, we generate $k$ explanation maps for the classifier's decision to classify $x$ as class $c$ using $k$ explanation techniques. Second, for each explanation map $h_j(x)$, we use the corresponding detector model to determine if the explanation map is normal or abnormal. 
For \sysname-CNN, each detector model directly provides a classification of normal ($g_{c,j}(h_j(x))$=0) or abnormal ($g_{c,j}(h_j(x))$=1).
On the other hand, for \sysname-AE, we obtain the reconstruction error for each explanation map $h_j(x)$ using the corresponding autoencoder $ae_{c,j}$. If the reconstruction error computed by a detector model is above its threshold, then it considers the explanation map to be abnormal. Finally, in both settings, if any of the $k$ detector models classifies the respective explanation map as abnormal, we infer that the input $x$ is an adversarial example.
\section{Experiments}
\label{sec_experiment}
In this section, we evaluate the effectiveness of \sysname. This section is organized as follows. 
First, we provide details on the experiment settings in Section \ref{experiment_settings}. 
In Section \ref{subsec:perf_normal}, we report the performance of the system on normal examples.
Then, in Section \ref{eval_blackbox}, we show the performance of \sysname{} on blackbox attacks. 
Next, we investigate the generalizability of \sysname-CNN in Section \ref{experiment_train_limited}.
We compare the performance of our approach with three state-of-the-art detection methods in Section \ref{baseline}. 
Finally, in Section \ref{adaptive_adversaries}, we present our evaluation on whitebox attacks. 

\begin{table}[t]
  \centering
  \begin{small}
  \caption{Evaluation of blackbox attacks.}
  \label{tab:eval_of_blackbox_attacks}
  \scalebox{0.95}{
  \begin{tabular}{P{0.05cm}P{0.4cm}P{0.60cm}P{1.5cm}P{0.5cm}P{0.75cm}P{1.1cm}P{1cm}}
  
    \toprule
     &\multicolumn{2}{c}{Attack} &Parameter &Cost (s) &Success \newline{} Rate &Prediction \newline Confidence &$L_2$ \newline Distortion\\
      
      \cmidrule{1-8}
      \multirow{7}{*}{\rotatebox[origin=c]{90}{MNIST}} &\multirow{3}{*}{$L_{\infty}$} 
        &CW$_{\infty}$ &- &90.57 &100\% &39.11\% &3.47 \\
      & &BIM &eps:0.3 &0.003 &99\% &99.94\% &4.10 \\
      & &MIM &eps:0.3 &0.003 &100\% &99.99\% &5.98 \\
      
      \cmidrule{2-8}
      &\multirow{1}{*}{$L_2$} &CW$_2$ &confidence:0 &0.001 &100\% &97.11\% &4.33 \\
      
      \cmidrule{2-8}
      &\multirow{2}{*}{$L_0$} 
        &CW$_0$ &- &8.98 &100\% &38.18\% &5.47 \\
      & &JSMA &gamma:0.2 &0.84 &94\% &76.86\% &7.00 \\
      \midrule
      \multirow{7}{*}{\rotatebox[origin=c]{90}{FMNIST}} &\multirow{3}{*}{$L_{\infty}$} 
        &CW$_{\infty}$ &- &90.07 &100\% &38.51\% &0.729 \\
      & &BIM &eps:0.3 &0.004 &98\% &100\% &4.06 \\
      & &MIM &eps:0.3 &0.004 &100\% &100\% &5.87 \\
      
      \cmidrule{2-8}
      &\multirow{1}{*}{$L_2$} 
        &CW$_2$ &confidence:0 &0.006 &100\% &96.09\% &2.20 \\
      
      \cmidrule{2-8}
      &\multirow{2}{*}{$L_0$} 
        &CW$_0$ &- &8.95 &100\% &37.80\% &2.88 \\
      & &JSMA &gamma:0.2 &0.96 &85\% &83.20\% &4.53 \\
      \midrule
      
      \multirow{7}{*}{\rotatebox[origin=c]{90}{CIFAR-10}} &\multirow{3}{*}{$L_{\infty}$} 
        &CW$_{\infty}$ &- &68.90 &100\% &26.91\% &1.19 \\
      & &BIM &eps:0.3 &0.008 &100\% &100\% &6.1 \\
      & &MIM &eps:0.3 &0.001 &100\% &100\% &7.9 \\
      
      \cmidrule{2-8}
      &\multirow{1}{*}{$L_2$} &CW$_2$ &confidence:0 &0.005 &100\% &94.52\% &3.86 \\
      
      \cmidrule{2-8}
      &\multirow{2}{*}{$L_0$} 
        &CW$_0$ &- &13.35 &100\% &27.09\% &2.98 \\
      & &JSMA &gamma:0.2 &6.42 &100\% &43.35\% &1.78 \\
    \bottomrule
  \end{tabular}
  }
  \end{small}
\end{table}
\begin{table}[t]
  \centering
  \caption{Classification accuracy on normal examples with and without defense.}
  \label{tab:classifier_acc_on_normal_wit_wo_def}
  \scalebox{0.85}{
  \begin{tabular}{P{1.34cm}P{1cm}P{1.17cm}P{1cm}P{1cm}P{0.9cm}P{1cm}}
    \toprule
    Dataset &Accuracy without \sysname{} &Top-1 Mean \newline Confidence &Accuracy with \sysname{} (CNN) &FP rate with \sysname{} (CNN) &Accuracy with \sysname{} (AE) &FP rate with \sysname{} (AE) \\
    \midrule
     MNIST &99.15\% &99.86\% &98.26\% &0.90\% &98.54\% &0.62\%\\
     FMNIST &90.68\% &97.86\% &89.70\% &1.08\% &89.91\% &0.85\%\\
     CIFAR-10 &84.54\% &76.64\% &83.74\% &0.95\% &83.85\% &0.82\%\\
    \bottomrule
  \end{tabular}
  }
  \vspace{-3mm}
  \end{table}

\subsection{Experimental settings}
\label{experiment_settings}
\textbf{Environment}. 
We implement the proposed framework using the Python libraries Keras and TensorFlow. We conducted our experiments on a Linux server with one GPU (GeForce RTX 2080 Ti) and CPU (Intel Xeon Silver 4116 processor).

\textbf{Image Datasets}. 
We evaluated the performance of our detection mechanism on three image datasets: MNIST \cite{backprop}, Fashion-MNIST (FMNIST) \cite{fmnist} and CIFAR-10 \cite{krizhevsky2009learning}. MNIST is a well-known gray-scale image dataset of handwritten digits from 0 to 9. FMNIST is a relatively more challenging dataset of article images where each example is a 28x28 grayscale image associated with a label from 10 classes (shirts, sandals, etc.). Both datasets consist of 60000 examples in the training set and 10000 examples in the testing set. CIFAR-10 is a colored image dataset of tiny 32x32x3 images used for object recognition. It comprises of 50000 training images and 10000 testing images. We chose MNIST and CIFAR-10 datasets as they are most widely used for evaluating defenses against adversarial attacks \cite{Ma-etal19NIC, magnet, Xu-etal17feature, Carlini-Wagner17notEasilyDetected}, and additionally used FMNIST as it provides more challenges for a gray-scale dataset.

\textbf{Training the Target Models}.
On MNIST and FMNIST datasets, we trained a CNN based target model with 54000 examples in the training set and 6000 examples in the validation set. For CIFAR-10, we trained the CNN based target model with 44000 examples in the training set and 6000 examples in the validation set.
For reproducibility, we refer to Appendix \ref{ap_train_target} where Table \ref{tab:classifier_arch} shows the CNN architectures, and Table \ref{tab:classifier_params} shows the hyperparameters for training the three target models. 

\textbf{Generating Adversarial Examples}.
As described in Section \ref{subsec_existing_attacks}, we generate adversarial examples using six state-of-the-art attacks- 
JSMA~\cite{papernot2016limitations}, BIM~\cite{bim}, MIM~\cite{dong2018boosting}, and CW$_0$, CW$_2$, and CW$_{\infty}$ variants of the CW attack~\cite{carlini2017towards}.
For JSMA, BIM, MIM, and CW$_2$ attacks, we created adversarial samples using their implementations in the Cleverhans library \cite{cleverhans}. For CW$_0$ and CW$_{\infty}$ attacks, we use the implementation from the authors \cite{carlini2017towards,nn_robust_attacks}.
For our evaluation, we adopt the \textit{target-next} attack setting in which the targeted label is the class next to the ground truth class modulo the number of classes (e.g., misclassify an input of class 4 to class 5). Table \ref{tab:eval_of_blackbox_attacks} shows a summary of our evaluation of the six blackbox attacks.

We generate adversarial examples for two purposes. First, as discussed in section \ref{subsec_detector}, we need adversarial examples to derive abnormal explanations for training and validating \sysname-CNN. To this end, we consider each class in a dataset and generate as many adversarial examples as the number of normal examples of that class in the training and validation sets. These adversarial examples are unevenly distributed by attack methods, due to relatively higher cost involved in conducting certain attacks. Column 5 in Table \ref{tab:eval_of_blackbox_attacks} shows the average cost (in seconds) to generate one adversarial example for different attacks. We observe that the CW$_{\infty}$, CW$_0$, and JSMA attacks incur much more overhead than remaining three attacks. Therefore, we generate 80\% of the examples using BIM, MIM, and CW$_2$ attacks, and the remaining using CW$_{\infty}$, CW$_0$, and JSMA attacks. We empirically found this distribution to be sufficient, based on performance on the validation sets.
Furthermore, seed images for these adversarial examples are randomly selected from normal examples of the source class. Note that we only utilize examples from the training (resp., validation) set towards training (resp., validating) \sysname; any example in the test set is considered non-accessible for this purpose, as is standard practice.

The second purpose is to evaluate the detection rate of \sysname.
To this end, for every dataset, we generate 100 adversarial examples using each attack. This number is consistent with previous works \cite{Ma-etal19NIC, Xu-etal17feature} and is limited since many attacks are too expensive to execute. In this process, we create the same number of adversarial examples for every target class to ensure a balanced evaluation. Furthermore, seed images for adversarial examples are taken from correctly classified examples in the test set so that the normal counterparts are unseen by the target model, and the resulting abnormal explanations are unseen by \sysname-CNN. 
 
In Table \ref{tab:eval_of_blackbox_attacks}, column 6 shows the success rate achieved by the six blackbox attacks when \sysname{} is not included as a defense. 
We consider an attack to be successful if the target model predicts the targeted class. The resulting examples from such attacks are termed as \textit{successful adversarial examples}. 
We observe that most attacks are very effective against three target models. The BIM, MIM, and CW$_2$ attacks are particularly effective in generating high-confidence adversarial examples as shown in column 7.

\textbf{Training \sysname-CNN and \sysname-AE}.
We refer to Table \ref{tab:arch_intr_based_classifier} and Table \ref{tab:arch_hp_exad_ae} in Appendix \ref{ap_train_exp_based} for details of the CNN architectures and hyperparameters used for training the CNN-based and autoencoder-based detector models, respectively.
For each setting, we use the same architecture and hyperparameters for all three datasets as the performance on the respective validation sets was found acceptable. 
As discussed in Section \ref{subsec_detector}, for each target class, we train a detector model for every explanation technique. 
For training and validation, while \sysname-CNN uses both normal and abnormal explanations, \sysname-AE only uses normal explanations.
To obtain normal explanations for a class, we take all its normal examples in our training and validation sets and generate corresponding explanations. 
For abnormal explanations (to be used by \sysname-CNN), we generate explanations of the adversarial examples being classified as the target class using each explanation technique. 
As discussed earlier, for this purpose, we had generated as many adversarial examples as the number of normal examples of each class in the training and validation sets. 
This provides us with balanced training and validation sets of normal and abnormal explanations.
For \sysname-CNN, we label the normal and abnormal explanations as negative and positive class, respectively. We train the detector models on the training set, and use the validation set for tuning the hyperparameters.
For \sysname-AE, we exclude the abnormal explanations in both training and validation sets (so that they online consist of normal explanations). Then, we train the detector models on the training set, and use the validation set for setting the threshold $t_{\textnormal{re}}$ values. 
Also, for computing the reconstruction error (equation \ref{eqn:recon_er}), we empirically find it sufficient to use the $L_2$ norm. Furthermore, we selected the threshold $t_{\textnormal{re}}$ such that the false-positive rate for any detector model is at most 0.2\% on its validation set.

\textbf{Comparison.} We compare \sysname{} with three state-of-the-arts- MagNet \cite{magnet}, Feature Squeezing (FS) \cite{Xu-etal17feature}, and LID \cite{ma2018characterizing}. For their implementation, we use the respective GitHub repositories. We follow instructions in the repositories and papers to identify optimal configurations. Feature Squeezing, in particular, allows many configurations for its squeezers. Consistent with the author's work, we utilize the optimal join-detection setting with multiple squeezers.
For MNIST and FMNIST, we use the combination of a 1-bit depth squeezer with 2x2 median smoothing. For CIFAR-10, we use a 5-bit depth squeezer with 2x2 median smoothing and 13-3-2 non-local means filter. To ensure fair comparison, for all detection methods, we set thresholds such that the false-positive rate on the validation set is at most 0.2\% (same as \sysname).
Additionally, our comparison could not include NIC ~\cite{Ma-etal19NIC} as it is yet to be made open-source, and we were not successful in reproducing the system.

\begin{table*}[t]
  \centering
  \normalsize
  \caption{Detection rate of \sysname{} on blackbox attacks and comparison with state-of-the-art detection methods.}
  \label{tab:classifier_acc_on_adv}
  \scalebox{0.85}{
  \begin{tabular}{P{1cm}P{1cm}P{1cm}P{1.5cm}P{1.5cm}P{1.7cm}P{1.4cm}P{1.6cm}P{1cm}P{1.2cm}}
    \toprule
    Dataset &\multicolumn{2}{c}{Attack} &Parameter &No Defense &\textbf{\sysname{}-CNN} &\textbf{\sysname{}-AE} &MagNet \cite{magnet} &FS \cite{Xu-etal17feature} &LID \cite{ma2018characterizing}\\
    \cmidrule{1-10}
    \multirow{7}{*}{\rotatebox[origin=c]{90}{MNIST}} 
    &\multirow{3}{*}{$L_{\infty}$} 
        &CW$_{\infty}$ &- &0\% &100\% &100\% &96\% &100\% &92\%\\
        & &BIM &eps:0.3 &1\% &100\%  &100\% &100\% &97.98\% &97.98\%\\
        & &MIM &eps:0.3 &0\% &100\%  &100\% &100\% &98\% &99\%\\

    \cmidrule{2-10}
    &\multirow{1}{*}{$L_2$} 
        &CW$_2$ &confidence:0  &0\% &100\%  &100\% &86\% &100\% &91\%\\
    
    \cmidrule{2-10}
    &\multirow{1}{*}{$L_0$} 
        &CW$_0$ &- &0\% &100\% &100\% &86\% &91\% &91\%\\
        & &JSMA &gamma:0.2 &6\% &100\% &100\% &84.04\% &100\% &93.62\%\\

    \midrule
    \multirow{7}{*}{\rotatebox[origin=c]{90}{FMNIST}}
    &\multirow{3}{*}{$L_{\infty}$} 
        &CW$_{\infty}$ &- &0\% &100\% &100\% &97\% &100\% &94\%\\
        & &BIM &eps:0.3 &2\% &100\%  &100\% &100\% &97.96\% &93.88\%\\
        & &MIM &eps:0.3 &0\% &100\%  &100\% &99\% &99\% &95\%\\
    \cmidrule{2-10}
    &\multirow{1}{*}{$L_2$} 
        &CW$_2$ &confidence:0  &0\% &100\%  &100\% &85\% &100\% &92\%\\
     
    \cmidrule{2-10}
    &\multirow{1}{*}{$L_0$} 
        &CW$_0$ &- &0\% &100\% &100\% &85\% &90\% &91\% \\
        & &JSMA &gamma:0.2 &15\% &100\% &100\% &87.06\% &100\% &92.94\%\\ 
    
    \midrule
    \multirow{7}{*}{\rotatebox[origin=c]{90}{CIFAR-10}} 
    &\multirow{3}{*}{$L_{\infty}$} 
        &CW$_{\infty}$ &- &0\% &99\% &100\% &84\% &98\% &90\% \\
        & &BIM &eps:0.3 &0\% &100\%  &100\% &100\% &52\% &97\% \\
        & &MIM &eps:0.3 &0\% &100\%  &100\% &100\% &51\% &97\% \\
    
    \cmidrule{2-10}
    &\multirow{1}{*}{$L_2$} &CW$_2$ &confidence:0  &0\% &100\%  &100\% &92\% &100\% &89\% \\
    
    \cmidrule{2-10}
    &\multirow{2}{*}{$L_0$} 
        &CW$_0$ &- &0\% &98\% &100\% &76\% &98\% &91\% \\
        & &JSMA &gamma:0.2 &0\%  &100\% &99\% &95\% &83\% &92\%\\
    \bottomrule
  \end{tabular}}
\end{table*}

\subsection{Performance on normal examples}
\label{subsec:perf_normal}
Table \ref{tab:classifier_acc_on_normal_wit_wo_def} shows the classification accuracy of the three target models on their test set (of normal examples). 
Without \sysname, we achieve an accuracy of 99.15\%, 90.68\%, and 84.54\% for MNIST, FMNIST, and CIFAR-10 datasets, respectively.
When \sysname{} is included as a defense, it is possible that a correctly classified normal example (by the target model) is misclassified as an adversarial example, termed as a \textit{false-positive} (FP). 
With \sysname-CNN, we obtained a false-positive rate of 0.90\% on MNIST dataset, as 89 of 9915 correctly classified normal examples are classified as adversarial. Thus, with \sysname-CNN, the accuracy of the target system is reduced only slightly to 98.26\%. Similarly, on FMNIST and CIFAR-10 datasets, the accuracy has a minor drop to 89.70\% and 83.74\%, respectively. 
The low false-positive rates are indicative of explanation maps of normal examples rarely being mistaken as abnormal by the detector models. This allows us to maintain a strict policy of classifying a test input as adversarial if even a single detector model considers its explanation map as abnormal.

With \sysname-AE, we obtain false-positive rates of 0.62\%, 0.85\%, and 0.82\% on the test sets of MNIST, FMNIST, and CIFAR-10 datasets, respectively. The accuracy of the target systems under this setting is 98.54\% for MNIST, 89.91\% for FMNIST, and 83.85\% for CIFAR-10 dataset. These results are close to the performance of \sysname-CNN on normal examples. In the following section, we show that both settings of \sysname{} can effectively detect adversarial attacks while maintaining these low false-positive rates.

\subsection{Evaluation on blackbox attacks}
\label{eval_blackbox}
Table \ref{tab:classifier_acc_on_adv} shows the detection rates of our approach on the six blackbox attacks. Columns 1-4 show datasets and details of the attacks. Column 5 shows the detection rate of adversarial examples when \sysname{} is not included as a defense (which corresponds to the success rate achieved by blackbox attacks on the target models). 
Columns 6 and 7 show the detection rates of \sysname-CNN and \sysname-AE, respectively. Note that, except when noted explicitly, ``detection rate" of a detection method refers to its detection rate on successful adversarial examples, consistent with previous work~\cite{Xu-etal17feature}. The remaining columns report our comparison with three state-of-the-art detectors, which we will discuss in Section \ref{baseline}.

We first consider the \sysname-CNN setting. For this setting, our approach achieves consistently high detection rates for all attacks across the three datasets. As shown in Table \ref{tab:classifier_acc_on_adv}, for MNIST and FMNIST datasets, we get 100\% detection rates for all six attacks.
For CIFAR-10 dataset, we obtain 98\% detection rate for CW$_2$ and CW$_0$ attacks, and 100\% detection rate for remaining attacks. 
For the \sysname-AE setting, the detection rate of adversarial examples is again consistently high. We achieve 100\% detection rate for all six attacks on MNIST and FMNIST datasets. On CIFAR-10 dataset, we obtain a 99\% detection rate for JSMA attack. For all other attacks on CIFAR-10 dataset, the detection rate is 100\%.

While we find both settings of \sysname{} are effective against adversarial attacks, each has its own relative advantages and disadvantages. A benefit of \sysname-AE is that it does not rely on adversarial examples for training or validation. This reduces the training overhead as many adversarial attacks incur significant cost (Table \ref{tab:eval_of_blackbox_attacks}). More importantly, being attack-independent, the performance of \sysname-AE indicates that our approach generalizes well to unknown attacks. 
But, compared to the effort required in setting the threshold $t_{\textnormal{re}}$ values for \sysname-AE, it is relatively simpler to tune the hyperparameters for \sysname-CNN. 
However, training the detector models in \sysname-CNN requires adversarial examples (to derive abnormal explanations). In the following section, we investigate the extent to which this requirement impacts the generalizability of \sysname-CNN in detecting unknown attacks.
 
\subsection{Generalizability of \sysname-CNN}
\label{experiment_train_limited}


\begin{table}[t]
  \centering
  \normalsize
  \caption{Detection rate of \sysname-CNN with limited attacks used in training.}
  \scalebox{0.82}{
  \begin{tabular}{ccccccc}
    \toprule
    \multirow{2}{*}{Attack} &\multicolumn{3}{c}{Train on BIM, MIM, CW$_2$, JSMA} &\multicolumn{3}{c}{Train only on CW$_2$} \\
    \cmidrule{2-7}
    &MNIST &FMNIST &CIFAR-10 &MNIST &FMNIST &CIFAR-10\\
    
    \midrule
    CW$_{\infty}$ &100\% &100\% &92\% &99\% &99\% &96\%\\
    BIM &100\% &100\% &100\% &97.98 \% &96.94\% &100\%\\
    MIM &100\% &100\% &100\% &89\% &86\% &100\%\\
    \midrule   
    CW$_2$ &100\% &100\% &100\% &100\% &100\% &100\%\\
    \midrule
    CW$_0$ &100\% &100\% &92\% &100\% &97\% &91\%\\
    JSMA &100\% &100\% &100\% &91.49\% &94.12\% &87\%\\
    \bottomrule
  \end{tabular}
  }
  \label{tab:classifier_acc_on_adv_wo_cw}
  \vspace{-3mm}
 \end{table}

\begin{table}[t]
  \centering
  \normalsize
  \caption{False-positive rates obtained for MagNet~\cite{magnet}, FS~\cite{Xu-etal17feature}, and LID~\cite{ma2018characterizing}.}
  \label{tab:fp_comparative}
 \scalebox{0.9}{
  \begin{tabular}{cccc}
    \toprule
    Dataset &MagNet~\cite{magnet} &FS~\cite{Xu-etal17feature} &LID~\cite{ma2018characterizing}\\
    \midrule
     MNIST &0.50\% &3.85\% &4.24\%\\
     FMNIST &0.81\% &3.76\% &3.89\%\\
     CIFAR-10 &4.25\% &4.81\% &5.36\%\\
    \bottomrule
  \end{tabular}}
  \vspace{-4mm}
\end{table}

In the generation of abnormal explanations, we notice that adversarial examples created using attacks of the same category ($L_\infty$ or $L_0$) produce similar explanations. 
We illustrate this phenomenon for interested readers in Figure \ref{fig:diff_attack_similar_exp} in Appendix \ref{ap_generalizability}. 
Using this observation, we leave out the CW$_\infty$ and CW$_0$ attacks, and only use adversarial examples from other four attacks for training \sysname-CNN. We keep other training and attack parameters same as before.
In Table \ref{tab:classifier_acc_on_adv_wo_cw}, columns 2-4 show the new performance of \sysname-CNN. On MNIST and FMNIST datasets, we find that \sysname-CNN effectively detects the unknown attacks. 
On CIFAR-10, \sysname-CNN still achieves a good detection rate of 92\% for both unknown attacks, CW$_\infty$ and CW$_0$.
We also observe that the detection of other attacks is not affected on any dataset. 
These results appear to indicate that abnormal explanations are influenced more by the original class of the seed images and the category (norm) of the attack, rather than the attack variant in that category used to craft the adversarial examples. This allows \sysname-CNN to generalize well to unknown attacks when we train on representative attacks from different categories. Furthermore, to study the generalizability in a more restrictive case, we only train \sysname-CNN on CW$_2$ attack. Columns 5-7 in Table \ref{tab:classifier_acc_on_adv_wo_cw} show the performance under this scenario. For all datasets, we find the results are good on the three CW attacks, with the lowest detection rate of 91\% obtained for CW$_0$ attack on CIFAR-10 dataset. For BIM attack, the performance remains consistently high across datasets. However, the detection is less effective for MIM attack on MNIST and FMNIST datasets, for which we have 89\% and 86\% detection rates, respectively, and for JSMA attack on CIFAR-10, for which we obtain 87\% detection rate. Overall, we see that \sysname-CNN can detect many unknown attacks even in this restrictive case, but there is still room for improvement. For boosting the detection in such scenarios, we can consider performing join-detection using both \sysname-CNN and \sysname-AE (which obtained high detection rates while being attack-independent). Building such joint-detectors to improve generalizability will be an interesting topic for future work.

\subsection{Comparison}
\label{baseline}
Table \ref{tab:classifier_acc_on_adv} shows the comparison of \sysname{} with three state-of-the-art adversarial detection methods- MagNet \cite{magnet}, Feature Squeezing \cite{Xu-etal17feature}, and LID \cite{ma2018characterizing}. The false-positive rates for these methods on the test sets are reported in Table \ref{tab:fp_comparative}.

\textbf{MagNet.} We find that MagNet's false-positive rates on the two grayscale datasets are marginally lower than those of \sysname-AE. However, it has much higher false-positive rate of 4.25\% on CIFAR-10 dataset.
We also find its detection performance to vary depending upon the dataset and attack-norm. MagNet uses trained autoencoders to detect adversarial examples, and to reform them based on the differences between the manifolds of normal and adversarial examples. MagNet's denoising strategy is quite effective against $L_\infty$ attacks, for which adversarial examples tend to have a large number of modified pixels, with a limit on the change per pixel. MagNet achieves high detection rates (most being 100\%) for $L_\infty$ attacks on both grayscale datasets. On CIFAR-10, a colored dataset, it achieves similar performance on BIM and MIM attacks, but relatively low detection rate of 84\% on CW$_\infty$ attack. Furthermore, we observe that MagNet's denoising mechanism is not as effective on $L_0$ attacks. These attacks make changes of high magnitude to very few pixels, thereby making denoising difficult. This is consistent with findings by Ma et al.~\cite{Ma-etal19NIC}. Similarly, we find MagNet's performance on CW$_2$ attack is not as good on MNIST and FMNIST datasets. Moreover, MagNet requires training a single detector network, which is computationally expensive. A benefit of our approach is that we use small detector models for every class, which are much easier to train. This makes our approach more practical.

\textbf{LID.} We find the detection performance of LID to be consistent across attacks and datasets. This method computes an LID value that captures the intrinsic dimensional properties of adversarial regions\cite{ma2018characterizing}. LID achieves its highest detection for $L_\infty$ attacks on the three datasets. For most of the other attacks, its detection was consistently above 90\%. However, as shown in Table \ref{tab:fp_comparative}, a downside of using LID values is the relatively higher false-positive rates on normal examples, which impacts the reliability of its classifications.

\textbf{Feature Squeezing.} For Feature Squeezing, we observe that joint-detection provides fairly consistent detection rates (Table \ref{tab:classifier_acc_on_adv}), but introduces high false-positive rates between 3.76\% to 4.81\% (Table \ref{tab:fp_comparative}). This is natural, given the use of a single threshold across all squeezers, consistent with original work~\cite{Xu-etal17feature}. Nevertheless, as reported by the authors, this can be improved by combining multiple squeezers with different thresholds in future work. 
Feature Squeezing obtains very high detection rates on all attacks on MNIST and FMNIST datasets. On CW$_2$ attack, it achieves 100\% detection rate on all three datasets. However, for $L_\infty$ attacks, while it obtains detection rates of above 98\% on CW$_\infty$ attack, we find it less effective against BIM and MIM attacks on CIFAR-10 dataset with a detection rate of nearly 52\%. This reflects upon the generalizability challenges in building squeezers. In contrast, our approach is more general and achieves consistent detection and false-positive rates. 

\subsection{Evaluation with adaptive adversaries}
\label{adaptive_adversaries}

\begin{table}
  \centering
  \caption{Evaluation of whitebox attacks.}
  \label{tab:eval_of_whitebox_attacks}
  \scalebox{0.80}{
  \begin{tabular}{P{1cm}P{1.6cm}P{2cm}P{1cm}P{1cm}}
    \toprule
     &Target Explanation Method &Mean Success Rate without defense &Cost (s) &$L_2$ \newline Distortion\\
     \midrule
      \multirow{5}{*}{\rotatebox[origin=c]{90}{MNIST}} 
        &LRP &99.00\% &137.07 &2.49\\
        &GBP &95.17\% &50.15 &2.64\\
        &IG &82.00\% &506.75 &2.81\\
        &PA &96.17\% &57.74 &2.55\\
        &GTI &91.50\% &48.96 &2.50\\
     \midrule
      \multirow{5}{*}{\rotatebox[origin=c]{90}{FMNIST}} 
        &LRP &99.17\% &169.36 &2.28\\
        &GBP &95.33\% &51.03 &2.34\\
        &IG &89.00\% &510.47 &2.33\\
        &PA &93.67\% &57.77 &2.76\\
        &GTI &91.50\% &49.12 &2.31\\
     \midrule
      \multirow{5}{*}{\rotatebox[origin=c]{90}{CIFAR-10}} 
        &LRP &99.00\% &138.83 &2.26\\
        &GBP &97.33\% &51.35 &2.24\\
        &IG &94.00\% &484.02 &3.87\\
        &PA &96.00\% &66.53 &3.20\\
        &GTI &95.33\% &66.80 &3.45\\

    \bottomrule
  \end{tabular}}
  \vspace{-3mm}
\end{table}

\begin{figure}[t]
  \centering
  \includegraphics[width=0.80\linewidth]{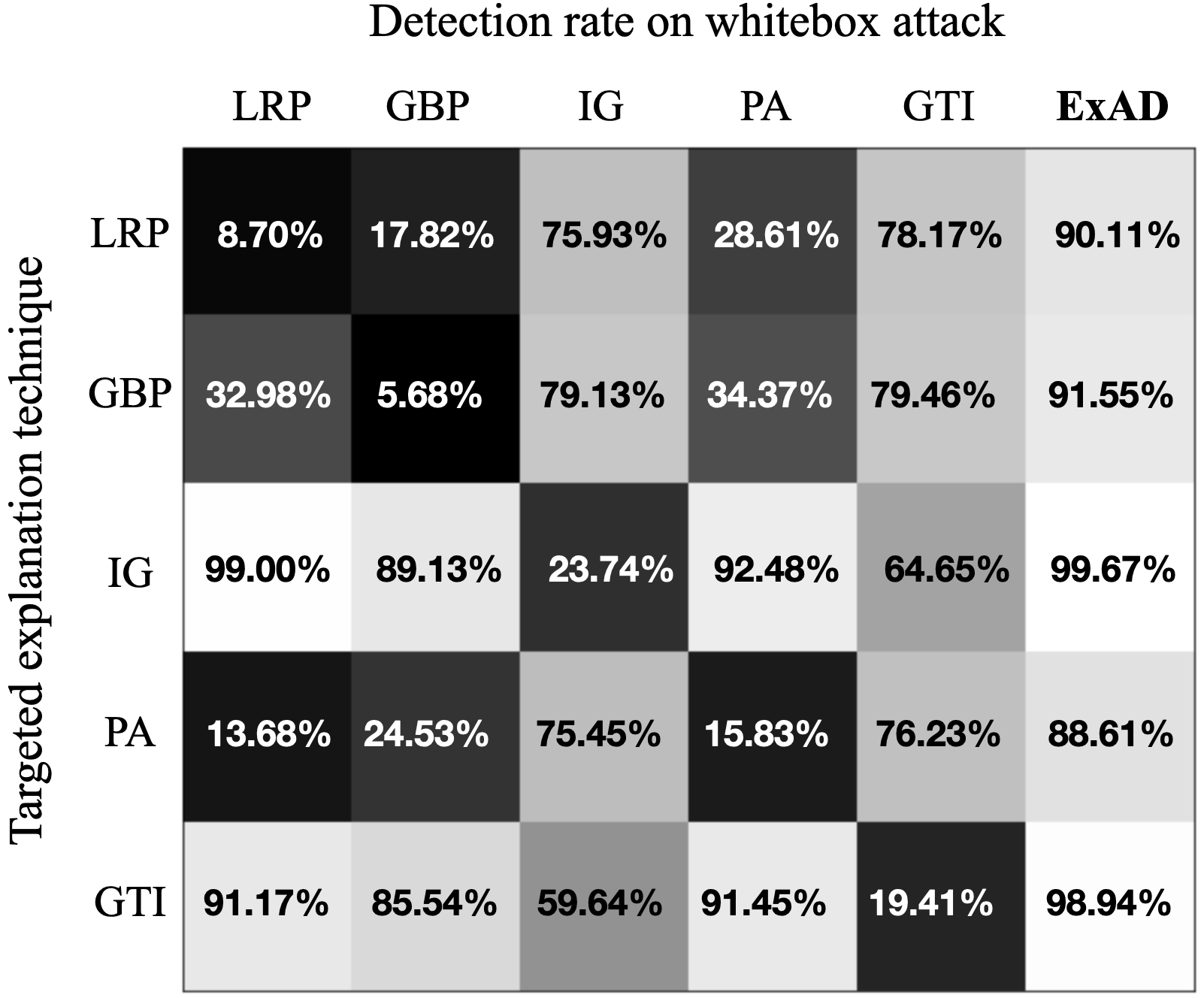}
  \caption{\textbf{Transferability of whitebox attack}}
  \label{fig:transferability_all}
  \vspace{-5mm}
\end{figure}

In this section, we evaluate our defense, with the \sysname-CNN setting, under whitebox threat model for an adaptive adversary. Here, we build upon recent research that shows that explanations can be unreliable \cite{kindermans2017reliability} and can be manipulated to produce a target explanation map \cite{zhang2020interpretable, dombrowski2019explanations}. Below, we present our approach to conduct a whitebox attack for generating an adversarial example.

\textbf{Whitebox Attack Approach.}
Given a normal or seed image $x \in \Bbb R^d$ with correctly classified class $C(x) = c$ by target model $f(\cdot)$, we follow a two-step process towards conducting a whitebox attack. First, we use a blackbox attack to generate an adversarial example $x'$ which is misclassified as $C(x')=t$, where $t$ is the targeted class. While $x'$ is likely to fool $f(\cdot)$, it is likely to be correctly classified as an adversarial example by the defense, which has not yet been accounted for by the attack. As a next step, we consider a target explanation technique $h_j$, which produces an explanation map $h_j(x')$ that is correctly classified as abnormal by the corresponding detector model $g_{t,j}(\cdot)$. Our goal in this step is to manipulate $x'$ to create a final adversarial example $x'' = x' + \delta x'$, such that
\begin{itemize}
    \item The target model's classification remains approximately constant, i.e. $f(x'') \approx f(x')$
    \item The explanation map $h_j(x'')$ is close to a target explanation map $h_j^t$ that is classified as normal by $g_{t,j}(\cdot)$
    \item The norm of the perturbation $\delta x'$ added is small so that it remains imperceptible. 
\end{itemize}
 To obtain the target explanation map $h_j^t$, we randomly select a normal example $x_r$ from class $t$ and check if its explanation map $h_j(x_r)$ is classified as normal by $g_{t,j}(\cdot)$. If so, we set the target explanation map as $h_j^t = h_j(x_r)$. Else, we repeat the process until we find such an example. This search is fairly quick because explanation maps of normal examples of a class are very likely to be correctly classified as normal by the detector models. Finally, we generate $x''$ by optimizing the following loss function
\begin{equation}
\label{wb_opt_eqn}
    \mathcal{L} = || h_j(x'') - h_j^t||^2 + \gamma||f(x'') - f(x')||^2
\end{equation}
with respect to $x''$ using gradient descent, such that $\|\delta x'\|_2 < \epsilon$. The first term in the loss function ensures that the explanation map for $x''$ is close to the target map, while the second term ensures the prediction by the target model is still the misclassified class $t$. The weighting of these two terms is controlled by hyperparameter $\gamma \in \Bbb R_+$. 
To compute the gradient with respect to the input $\triangledown h_j(x'')$, we follow the strategy by Dombrowski et al. of replacing relu with the softplus function to circumvent the problem of vanishing second-derivative \cite{dombrowski2019explanations}. A similar strategy of approximating relu was used by Zhang et al.~\cite{zhang2020interpretable}.
After the optimization completes, we test whether the manipulated image $x''$ fools the original relu-based target model $f(\cdot)$ as well as our defense. We provide an illustration of our whitebox approach in Appendix \ref{ap_illustration_whitebox}.

\textbf{Evaluation of Whitebox Attack.}
To perform the first step of the above approach, we re-use the successful adversarial examples $X'$ created using blackbox attacks. However, while targeting integrated gradients (IG), we only used adversarial examples from CW$_2$ attack as we find targeting IG to incur very high cost. For targeting remaining techniques, we use adversarial examples from all six attacks.
Then, for each adversarial example $x' \in X'$, we perform the second step using the optimization process described above to obtain final adversarial examples $X''$.

Table \ref{tab:eval_of_whitebox_attacks} shows a summary of the whitebox attack. Column 3 shows the mean success rate of the final adversarial examples in retaining the desired misclassification in the target model (when \sysname{} is not included). The mean is computed over success rates obtained for different attacks (except for IG where we only use CW$_2$ attack). We do not show individual success rates for each attack as they were very similar for any target technique.
In Table \ref{tab:eval_of_whitebox_attacks}, we observe that targeting LRP results in the highest success rate, with least $L_2$ distortion. However, as shown in Column 4, the average time required per example to target LRP is over 130 seconds which is quite high compared to targeting GBP, PA, or GTI techniques.

Figure \ref{fig:transferability_all} shows the results of the whitebox attack. Each row represents the targeted explanation technique. Columns 1-5 show the detection rate obtained by individual detector models corresponding to the five explanation techniques in correctly classifying explanation maps as abnormal. Column 6 shows the overall performance by \sysname-CNN{} in correctly classifying the examples as adversarial. 
From Figure \ref{fig:transferability_all}, we make several interesting observations. 
First, we notice the values along the diagonal (from top-left to bottom-right) are all very low. This is natural as the detector model corresponding to the targeted technique is expected to correctly classify very few explanations as abnormal. For instance, targeting GTI causes its detector model to have a detection rate of only 19.41\%. 
Second, we observe that targeting gradient-based techniques do not severely impact detector models of propagation-based techniques, and vice-versa. 
For instance, on targeting IG or GTI, the detector models corresponding to LRP, GBP, and PA still achieve detection rates above 85\%.
This is consistent with the transferability findings by Zhang et al.~\cite{zhang2020interpretable}. On a different set of diverse explanation techniques, the authors showed that manipulated images created by targeting one technique rarely produce desirable explanations (which are close to the target map) on other techniques.
Finally, we find that targeting propagation-based (resp., gradient-based) techniques transfer well to detector models corresponding to the other propagation-based (resp., gradient-based) techniques.  
For instance, targeting LRP results in the GBP-based detector model to have a detection rate of only 17.82\%. The same phenomenon can be observed for gradient-based techniques (IG and GTI). 
This empirically supports the need to have diversity in the explanation methods to build robustness against adaptive attacks.
As shown in Column 6 of Figure \ref{fig:transferability_all}, using an ensemble of gradient-based and propagation-based techniques, \sysname{} is able to significantly limit the success rate of whitebox attacks. 
We find \sysname{} to be relatively more robust when gradient-based techniques are targeted. We achieve 99.67\% and 98.94\% detection rate for IG and GTI as the target, respectively. For propagation-based techniques, the highest impact is caused by targeting PA, which results in a detection rate of 88.61\%. For the case of targeting LRP and GBP, we obtain detection rates of 90.11\% and 91.55\%, respectively. Appendix \ref{ap_transferability} includes further analysis on transferability of whitebox attack for individual datasets.
\section{Discussion}
\label{discussion}

\subsection{Fragility of Explanations}
\label{eval_reliability_of_intr}
In Section \ref{adaptive_adversaries}, we discussed the reliability of explanations in context of an adaptive adversary.
Recently, there has also been research that shows the fragility of explanations in an adversarial context. In this section, we discuss two related scenarios and any potential impact to our defense strategy.

\subsubsection{Hiding the attack from explanations}
Recently, \textit{adversarial patches}~\cite{brown2017adversarial,karmon2018lavan} were introduced to make adversarial examples more practical in the physical world. This attack restricts the spatial dimensions of the perturbation, but removes the imperceptibility constraint. However, Subramanya \textit{et al.} \cite{subramanya2018towards} demonstrated that we can generate adversarial patches that not only fool the prediction by the target classifier, but also change the explanation of the modified example such that the adversarial patch is no longer considered important. Nevertheless, here the attacker only manages to make the explanation technique focus in a region outside the adversarial patch; she does not try to make the explanation itself appear more normal for the target class. Therefore, even though such adversarial examples may evade the explanation mechanism, they are still likely to be correctly classified as adversarial by our defense. 

\subsubsection{Changing the explanation but not the classification}
\label{subsec_change_expl_not_classif}
An attacker may add adversarial perturbations which produce examples that are classified into the same class, but have very different explanations~\cite{ghorbani2019interpretation}.
With our defense, if such explanations are classified as abnormal by the detector models, then the corresponding inputs would be considered adversarial. 
Nevertheless, we believe the impact of this attack may depend upon the nature of the application where the defense is being used. 
If the perturbed examples resulting from such attacks are not considered normal for the system, then the abnormality produced in the explanations can be beneficial because the examples will likely be classified as adversarial. 
However, if an application considers such perturbed examples as normal, such as if the norm of perturbation is within an allowed threshold, then our defense could result in false-positives. We refer to Figure \ref{fig:acc_on_noisy} in Appendix \ref{ap_fragility} which shows the effect of such attacks on the target classifier's accuracy for this case.
For such applications, we would require explanation techniques to be robust enough to allowed perturbations. We leave further research towards such methods as future work.




\subsection{Limitations}
Our detection mechanism and scope of evaluation has certain limitations. 
First, our current evaluation only considers the targeted attack setting for generating adversarial examples. In future work, we will extend our evaluation to cover untargeted adversarial attacks as well. 
Second, 
currently we do not have a unified optimization-based approach that simultaneously (and successfully) targets multiple explanation techniques. We attempt doing so in two ways. One approach is to modify the loss function \ref{wb_opt_eqn} as follows
\begin{equation}
\label{wb_opt_eqn2}
    \mathcal{L} = \Big(\sum_{j=1}^{5}|| h_j(x'') - h^t||^2\Big) + \gamma||f(x'') - f(x')||^2
\vspace{-1mm}
\end{equation}
Here, we create the target map $h^t$ using any one of the explanation techniques. However, in this case, when the target map is created using a gradient-based technique, we did not notice any significant change in the performance of the detector models corresponding to propagation-based techniques, and vice-versa. The results remained consistent with our findings in Figure \ref{fig:transferability_all}. One reason behind this could be the diversity in explanation techniques due to which the target maps required by gradient-based techniques differ substantially from those required by propagation-based techniques. 
Then, we further consider modifying the loss function \ref{wb_opt_eqn2} as follows
\begin{equation}
    \mathcal{L} = \Big(\sum_{j=1}^{5}|| h_j(x'') - h_j^t||^2\Big) + \gamma||f(x'') - f(x')||^2
\vspace{-1mm}
\end{equation}
Here, target maps used are created using the corresponding explanation techniques. However, in this case, we found the explanation-loss component did not reduce much during the optimization process. Moreover, it often led to memory errors on the GPU. One reason for this is that in the current whitebox attack framework\cite{dombrowski2019explanations}, each explanation technique requires a different set of hyperparameters (e.g., learning rate, $\beta$ growth, and iterations), as shown in Table \ref{tab:hyperparameters_whitebox} in Appendix \ref{ap_whitebox}.
Therefore, we find that simultaneously attacking multiple explanation techniques is considerably difficult for an adaptive adversary. We leave further exploration on improving the efficiency and effectiveness of such attack to future work.

\section{Conclusion}
\label{sec_conclusion}
We proposed \sysname, a framework to detect adversarial examples using an ensemble of explanation techniques. The use of explanations is motivated by the distinguishability between normal and abnormal explanations for any target class. Furthermore, motivated by previous work on N-variant systems, we used an ensemble of gradient-based and propagation-based explanation techniques to introduce diversity in our defense. Experiments showed that our approach is effective against blackbox attacks. We also find that \sysname{} significantly limits the success rate of whitebox attacks. In this process, we made interesting findings on the transferability of adaptive attacks. We acknowledge the possibility of more sophisticated whitebox attacks in future, and hope our work will inspire further research in this direction. 
We believe our proposed defense can be used in conjunction with state-of-the-art detection methods to boost the detection of adversarial attacks. 
\bibliographystyle{ACM-Reference-Format}
\bibliography{reference.bib}

\pagebreak
\appendix
\section{Similarity in Normal Explanations}
\label{ap_similarity_normal}
\begin{figure}[h]
  \vspace{-2mm}
  \includegraphics[width=0.85\linewidth]{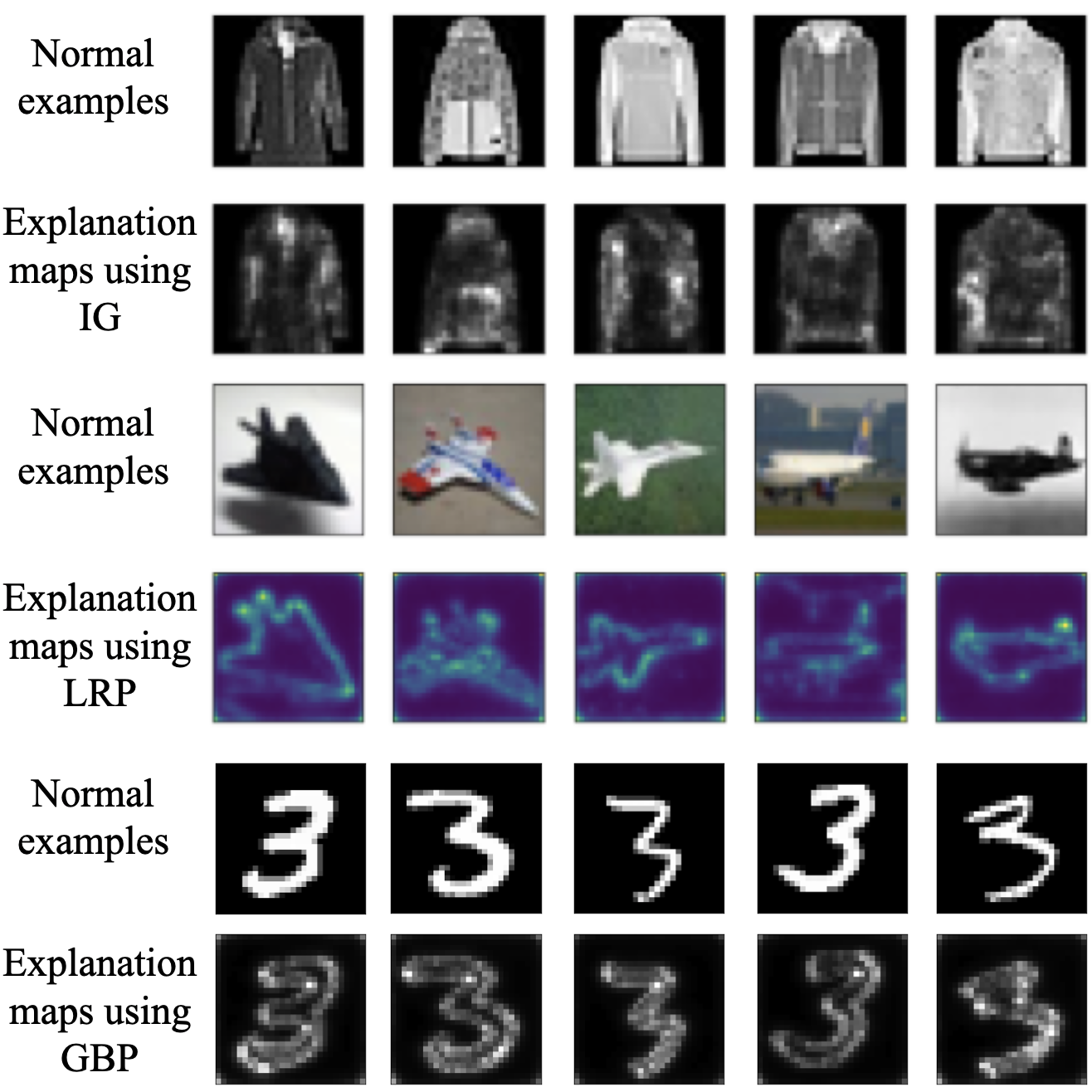}
  \caption{\textbf{Similarity in normal explanations.}}
  \label{fig:intr_of_gen_and_adv}
  \vspace{-3mm}
\end{figure}

Our approach is motivated by the observation that normal examples of a class tend to have similar (normal) explanations, and that a detector model can learn to distinguish between them and the (abnormal) explanations of adversarial examples targeting that class.
Figure \ref{fig:intr_of_gen_and_adv} shows an example of the similarity in normal explanations. The first row shows five normal examples from the Coat class of FMNIST dataset. The third row shows normal examples from the Airplane class of CIFAR-10 dataset. The fifth row shows normal examples of class Three from MNIST dataset. The second, fourth, and sixth rows show corresponding explanations for the preceeding row using the IG, LRP, and GBP techniques, respectively. 

\section{Implementation Details}
\subsection{Training the Target Models}
\label{ap_train_target}
Table \ref{tab:classifier_arch} shows the architecture of the target models for MNIST, FMNIST, and CIFAR-10 datasets.

\begin{table}[h]
  \centering
  \caption{Architecture of the image classifiers to be defended.}
  \label{tab:classifier_arch}
  \scalebox{0.9}{
  \begin{tabular}{P{1.2cm}P{1.1cm}P{1.3cm}P{0.8cm}P{1.3cm}P{0.8cm}}
    \toprule
    \multicolumn{2}{c}{MNIST} &\multicolumn{2}{c}{FMNIST} &\multicolumn{2}{c}{CIFAR-10}\\
    \midrule
     Conv.ReLU &8x8x64 &Conv.ReLU &3x3x32 &Conv.ReLU &3x3x64\\
     Conv.ReLU &6x6x128 &Conv.ReLU &3x3x32 &Conv.ReLU &3x3x128\\
     Conv.ReLU &5x5x128 &MaxPooling &2x2 &AvgPooling &2x2\\
     Softmax &10 &Conv.ReLU &3x3x64 &Conv.ReLU &3x3x128\\
      & &Conv.ReLU &3x3x64 &Conv.ReLU &3x3x256\\
      & &MaxPooling &2x2 &AvgPooling &2x2\\
      & &Dense.ReLU &200 &Conv.ReLU &3x3x256\\
      & &Dense.ReLU &200 &Conv.ReLU &3x3x512\\
      & &Softmax &10 &AvgPooling &2x2\\
      & & & &Conv.ReLU &3x3x10\\
      & & & &Softmax &10\\
    \bottomrule
  \end{tabular}
  }
\end{table}

Table \ref{tab:classifier_params} shows the training and architecture hyperparameters for the target models of the three datasets. 

\begin{table}[h]
  \centering
  \normalsize
  \caption{Training and architecture hyperparameters of the image classifiers to be defended}
  \label{tab:classifier_params}
  \begin{tabular}{cccc}
    \toprule
    Hyperparameter &MNIST &FMNIST &CIFAR-10\\
    \midrule
     Learning Rate &0.001 &0.01 &0.001\\
     Optimization Method &Adam &SGD &Adam\\
     Batch Size &128 &128 &256\\
     Epochs &50 &50 &50\\
     Padding (Conv layers) &Valid &Valid &Same\\
    \bottomrule
  \end{tabular}
\end{table}

\subsection{Training Detector Models of \sysname}
\label{ap_train_exp_based}
Table \ref{tab:arch_intr_based_classifier} shows the architecture and hyperparameters used for \sysname-CNN.

\begin{table}[h]
    \centering
    \normalsize
    \caption{Architecture and hyperparameters of \sysname-CNN}
    \scalebox{0.9}{
    \begin{tabular}{P{1.8cm}P{1.5cm}P{3cm}P{1cm}}
        \toprule
        \multicolumn{2}{c}{Architecture} &\multicolumn{2}{c}{Hyperparameters}\\
        \midrule
        Conv.ReLU &3x3x32 &Learning Rate &0.01\\
        Conv.ReLU &3x3x64 &Optimization Method &Adam\\
        MaxPooling &2x2 &Batch Size &32\\
        
        Conv.ReLU &3x3x128 &Epochs &50\\
        Conv.ReLU &3x3x128 &Padding (Conv layers) &Same\\
        MaxPooling &2x2\\
        
        Dense.ReLU &512\\
        Dense.ReLU &64\\
        Softmax &2\\
        \bottomrule
    \end{tabular}
    }
    \label{tab:arch_intr_based_classifier}
\end{table}

Table \ref{tab:arch_hp_exad_ae} shows the architecture and hyperparameters used for \sysname-AE. For MNIST and FMNIST datasets, $H=W=28$. For CIFAR-10 dataset, $H=W=32$.

\begin{table}[h]
    \centering
    \normalsize
    \caption{Architecture and hyperparameters of \sysname-AE}
    \scalebox{0.9}{
    \begin{tabular}{P{1.8cm}P{1.5cm}P{3cm}P{1cm}}
        \toprule
        \multicolumn{2}{c}{Architecture} &\multicolumn{2}{c}{Hyperparameters}\\
        \midrule
        Dense.ReLU &HxW &Learning Rate &$10^{-5}$\\
        Dense.ReLU &400 &Optimization Method &Adam\\
        Dense.ReLU &20 &Batch Size &32\\
        Dense.ReLU &400 &Epochs &100\\
        Dense.ReLU &HxW &\\
        Softmax &2\\
        \bottomrule
    \end{tabular}}
    \label{tab:arch_hp_exad_ae}
\end{table}

\subsection{Performing Whitebox Attack}
\label{ap_whitebox}
Table \ref{tab:hyperparameters_whitebox} shows the hyperparameters for the whitebox attack.

\begin{table}[h]
    \centering
    \normalsize
    \caption{Hyperparameters used in whitebox attack.}
    \scalebox{0.9}{
    \begin{tabular}{cccc}
        \toprule
        Method &Iterations &Learning Rate &Factors\\
        \midrule
        LRP &1500 &$10^{-3}$ &2x$10^{-4}$, $10^6$\\
        GBP &1500 &$10^{-3}$ &$10^{11}$, $10^6$\\
        IG &500 &$5x10^{-3}$ &$10^{11}$, $10^6$\\
        PA &1500 &$2x10^{-3}$ &$10^{11}$, $10^6$\\
        GradxInput &1500 &$10^{-3}$ &$10^{11}$, $10^6$\\
        \bottomrule
    \end{tabular}}
    \label{tab:hyperparameters_whitebox}
\end{table}

\section{Generalizability of \sysname-CNN}
\label{ap_generalizability}
\begin{figure}[h]
  \includegraphics[width=0.76\linewidth]{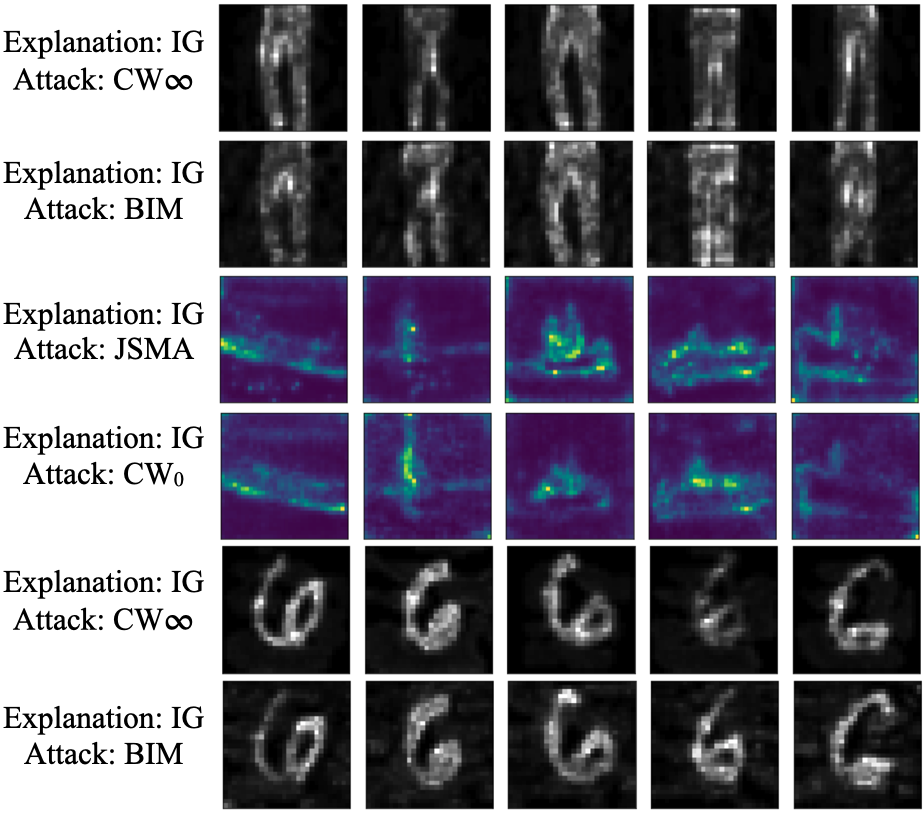}
  \caption{\textbf{Similarity in explanation maps of adversarial examples created using different attacks.}}
  \label{fig:diff_attack_similar_exp}
  \vspace{-1mm}
\end{figure}

Figure \ref{fig:diff_attack_similar_exp} shows explanation maps produced by the integrated gradients (IG) technique for adversarial examples created using different attacks. Rows 1 and 2 show explanation maps for adversarial examples which were created using CW$_\infty$ and BIM attacks, respectively. Both attacks come under the $L_\infty$ category. The adversarial examples are targeting the Pullover class and their seed images are taken from the Trouser class of FMNIST dataset. Comparing explanation maps in row 1 and row 2 shows that adversarial examples created using different attacks, under the same category ($L_\infty$ in this case) can result in similar explanation maps. This can also be observed for CIFAR-10 dataset (rows 3-4), where we use two $L_0$ attacks, and MNIST dataset (rows 5-6), where we again use two $L_\infty$ attacks.

\section{Illustration of Whitebox Attack}
\label{ap_illustration_whitebox}
\begin{figure}[t]
  \includegraphics[width=1\linewidth]{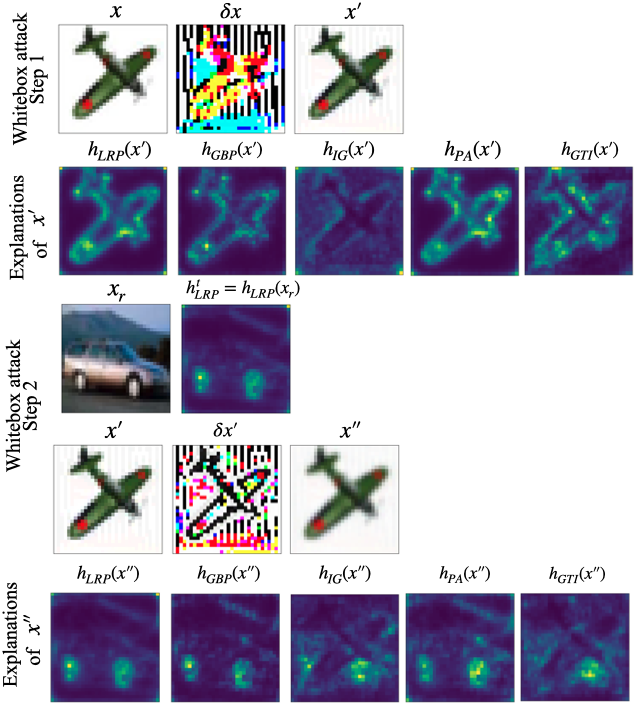}
  \caption{\textbf{Illustration of whitebox attack.}}
  \label{fig:whitebox_illustration}
\end{figure}

Figure \ref{fig:whitebox_illustration} shows an illustration of our whitebox attack approach, discussed in Section \ref{adaptive_adversaries}. The leftmost image in the first row shows a seed image $x$, which is a normal example from class Airplane of CIFAR-10 dataset. In the first step, we use CW$_\infty$ attack to add perturbations $\delta x$ to $x$, which results in the adversarial example $x' = x + \delta x$. The rightmost image of row 1 shows $x'$. The example $x'$ is misclassified as class Automobile by target model $f(\cdot)$. The second row shows the explanation maps produced for $x'$ by the five techniques. We find that all detector models classify these explanation maps as abnormal. We observe in row 2 that the explanation maps are not consistent with the expected normal explanations of class Automobile. As an adaptive adversary, we now intend to target an explanation technique. For this illustration, we choose to target LRP. To this end, we randomly select an example $x_r$ from class Automobile, for which the explanation map produced by LRP is classified as normal by the corresponding detector model. We show $x_r$ as the first image in row 3. Its explanation map by LRP, shown as the second image in row 2, is set as the target map $h^t_{LRP}$. Then, we perform the optimization for the loss function \ref{wb_opt_eqn} (shown in Section \ref{adaptive_adversaries}) using $x'$ and $h^t_{LRP}$. After the optimization completes, we obtain the final adversarial example $x'' = x' + \delta x'$, which is shown as the rightmost image in row 4. The bottom row shows the explanation maps produced by the five techniques for $x''$. We observe that targeting LRP results in the explanation map produced by LRP (first image in row 5) to be very close to the target map. Also, the explanation maps for GBP (second image in row 5) and PA (fourth image in row 5) are also fairly close to the target map. We find that all three of these explanation maps are classified as normal by their respective detector models. However, we observe that the explanation maps produced by the gradient-based techniques, i.e., IG (third image in row 5) and GTI (rightmost image in row 5), are not as close to the target map. Both these explanation maps are classified as abnormal by their respective detector models. Therefore, using an ensemble of detector models corresponding to diverse explanation techniques, our defense is able to mitigate this whitebox attack by correctly identifying $x''$ as an adversarial example.

\section{Transferability of Whitebox Attacks}
\label{ap_transferability}
\begin{figure*}[t]
  \includegraphics[width=1\textwidth]{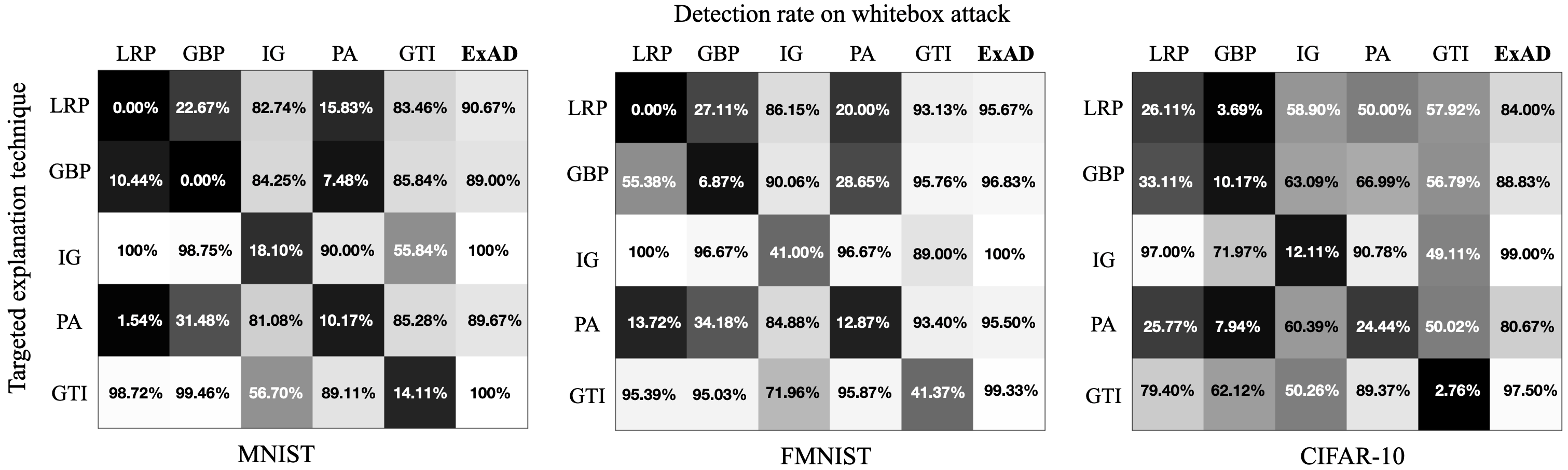}
  \caption{\textbf{Transferability of whitebox attack on individual datasets.}}
  \label{fig:whitebox_transferability_individual}
\end{figure*}

Figure \ref{fig:whitebox_transferability_individual} shows results on the transferability of whitebox attack on explanation-based detector models, as well as the detection rates obtained by our defense (under \sysname-CNN setting), for the three datasets. Previously, we had only shown mean values across datasets in Figure \ref{fig:transferability_all}. 
In Figure \ref{fig:whitebox_transferability_individual}, each value shows the mean detection rate for adversarial examples created using different attacks (except while targeting IG technique which only uses CW$_2$ attack) and considering all target classes. On all three datasets, we find that targeting a gradient-based technique has relatively less impact on detector models corresponding to propagation-based techniques. For instance, targeting GTI on MNIST causes the IG-based detector model to have a detection rate of only 56.70\% whereas detector models corresponding to LRP, GBP, and PA have detection rates above 89\%. Interestingly, on CIFAR-10 dataset, we observe that the detector models corresponding to IG and GTI are impacted by most techniques, although the impact is relatively higher when we target either of IG or GTI. For instance, on CIFAR-10 dataset, targeting PA has a noticeable impact on IG-based detector model as it achieves a detection rate of 60.39\%, whereas targeting GTI results in the IG-based detector model to have a relatively lower detection rate of 50.26\%.


In terms of impact on \sysname, on all three datasets, targeting gradient-based techniques (IG and GTI) is relatively less effective. For instance, on MNIST, we obtain 100\% detection rate by our approach while targeting IG and GTI techniques. In contrast, we observe that targeting propagation-based techniques is more effective across datasets. On MNIST dataset, the detection rate of the defense on propagation-based techniques is nearly 89\%. On CIFAR-10 dataset, targeting PA results in the lowest detection rate of 80.67\% by \sysname. On FMNIST dataset, the overall performance is fairly consistent while targeting propagation-based techniques. The highest impact is produced by targeting PA and LRP, for which \sysname{} obtains detection rates of 95.50\% and 95.67\%, respectively.

\section{Fragility of Explanations}
\label{ap_fragility}
\begin{figure}[h]
  \centering
  \includegraphics[width=1.0\linewidth, scale=0.5]{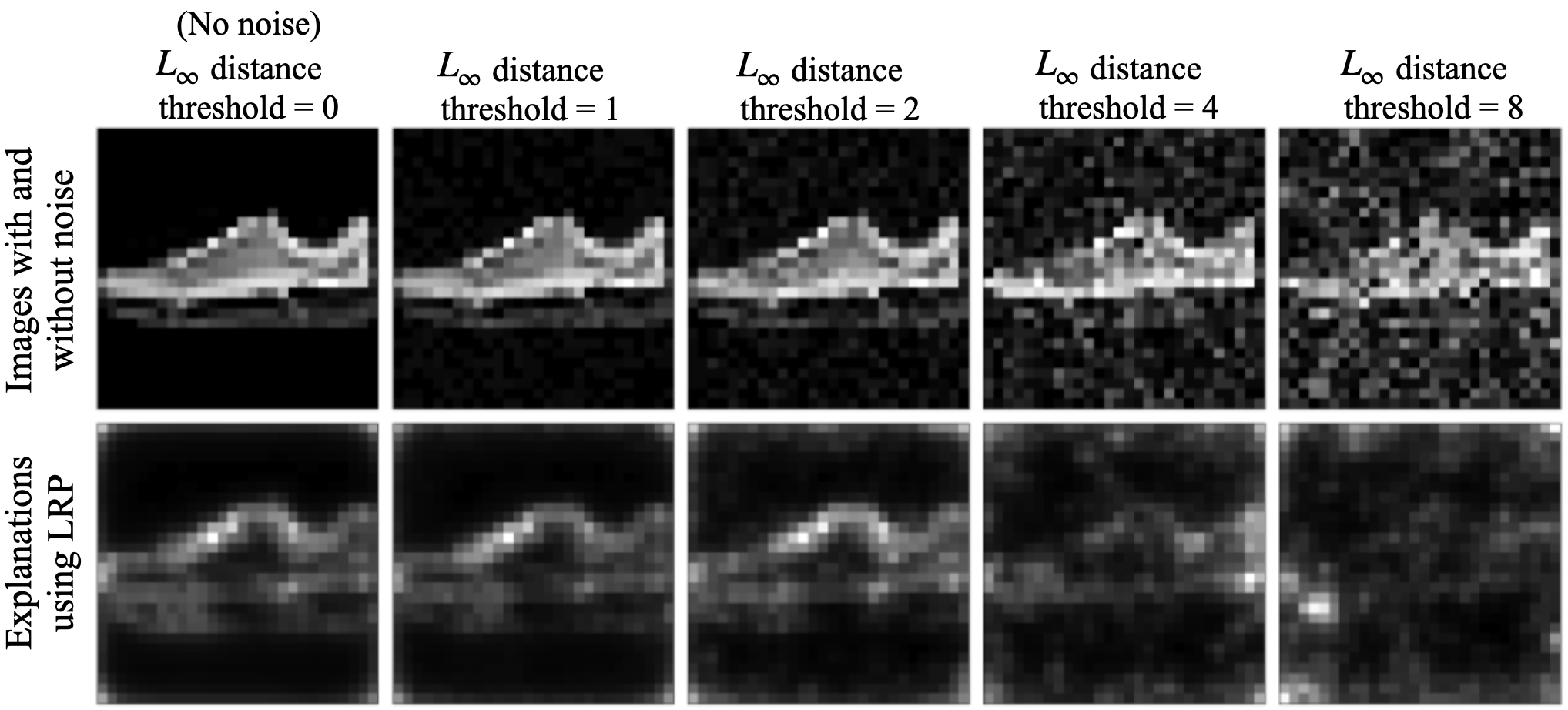}
  \caption{ Effect of adding random perturbations, with increasing threshold of $L_\infty$ distance, on explanations by LRP. 
  }
  \label{fig:intr_shirt_with_noise}
\end{figure}

\begin{figure}[h]
  \centering
  \includegraphics[width=0.65\linewidth, scale=0.5]{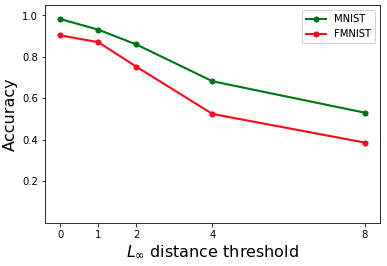}
  \caption{\textbf{Impact of fragility of explanations \cite{ghorbani2019interpretation}.}}
  \label{fig:acc_on_noisy}
\end{figure}

In Section \ref{subsec_change_expl_not_classif}, we discussed an attack which shows the fragility of explanations~\cite{ghorbani2019interpretation}. Figure \ref{fig:intr_shirt_with_noise} shows an example of this attack for the case of random perturbations. 
The leftmost image in first row is a normal example from the Sneaker class of FMNIST dataset. The next image in first row shows a manipulated image created by adding random perturbations to the normal example such that the prediction remains unchanged, and the perturbations do not exceed a threshold value of 1, in terms of $L_\infty$ distance. The next three images are created in a similar manner but with increased threshold values of 2, 4, and 8, respectively. The second row shows the corresponding explanations produced by LRP technique. We observe that with increased noise threshold, the explanations also become noisy.
With our defense, if such explanations are classified as abnormal, then the corresponding inputs would be considered adversarial (false-positive). Figure \ref{fig:acc_on_noisy} shows that adding random perturbations to normal examples results in a decline of the target classifier's classification accuracy with increasing noise threshold. We discussed in Section \ref{subsec_change_expl_not_classif} that the impact of this attack depends on the nature of the application using the defense. For instance, our approach will not adversely impact applications which consider such perturbed examples as abnormal.









\end{document}